\documentclass[letterpaper, 10 pt, conference]{ieeeconf}  %

\IEEEoverridecommandlockouts                              %

\usepackage{graphicx}

\usepackage{colortbl}
\usepackage{color}
\usepackage{array}
\usepackage{amsmath, amssymb, scalerel}
\usepackage{nccmath}
\usepackage{booktabs}
\usepackage{tabularx}
\usepackage{algpseudocode}
\usepackage[table,dvipsnames]{xcolor}
\usepackage{todonotes}
\usepackage{multicol,multirow}
\usepackage{rotating}
\let\labelindent\relax
\usepackage{enumitem,kantlipsum}

\usepackage{arydshln}
\usepackage{url}

\usepackage{transparent}

\DeclareMathOperator*{\concat}{\scalerel*{\Vert}{\sum}}
\DeclareMathOperator*{\argmax}{arg\,max}

\usepackage{xspace}
\makeatletter
\DeclareRobustCommand\onedot{\futurelet\@let@token\@onedot}
\def\@onedot{\ifx\@let@token.\else.\null\fi\xspace}
\def\eg{\emph{e.g}\onedot} 
\def\ie{\emph{i.e}\onedot} 
 
\def\etc{\emph{etc}\onedot}

\makeatother

\makeatletter
\def\adl@drawiv#1#2#3{%
        \hskip.5\tabcolsep
        \xleaders#3{#2.5\@tempdimb #1{1}#2.5\@tempdimb}%
                #2\z@ plus1fil minus1fil\relax
        \hskip.5\tabcolsep}
\newcommand{\cdashlinelr}[1]{%
  \noalign{\vskip\aboverulesep
           \global\let\@dashdrawstore\adl@draw
           \global\let\adl@draw\adl@drawiv}
  \cdashline{#1}
  \noalign{\global\let\adl@draw\@dashdrawstore
           \vskip\belowrulesep}}
\makeatother

\definecolor{LightCyan}{rgb}{0.88,1,1}
\definecolor{LightBlue}{rgb}{.867,.922,.969}
\definecolor{LightYellow}{rgb}{1,.949,.8}
\newcolumntype{a}{>{\columncolor{LightBlue}}c}

\usepackage{soul}
\sethlcolor{LightYellow}

\newcommand{\um}[1]{\textcolor{black}{#1}}

\newcommand{\rev}[1]{\textcolor{black}{#1}}

\title{\LARGE \bf
Online Continual Learning for Robust Indoor Object Recognition}

\author{Umberto Michieli and Mete Ozay%
\thanks{All authors are with Samsung Research UK.\protect\\
        {\tt\small \{n.surname\}@samsung.com}}%
}

\begin{document}

\maketitle
\thispagestyle{empty}
\pagestyle{empty}

\begin{abstract}
Vision systems mounted on home robots need to interact with unseen classes in changing environments. Robots have limited computational resources, labelled data and storage capability. These requirements pose some unique challenges: models should adapt without forgetting past knowledge in a data- and parameter-efficient way. We characterize the problem as few-shot (FS) online continual learning (OCL), where robotic agents learn from a non-repeated stream of few-shot data updating only a few model parameters. Additionally, such models experience variable conditions at test time, where objects may appear in different poses (\eg, horizontal or vertical) and environments (\eg, day or night). To improve robustness of CL agents, we propose RobOCLe, which; 1) constructs an enriched feature space computing high order statistical moments from the embedded features of samples; and 2) computes similarity between high order statistics of the samples on the enriched feature space, and predicts their class labels. We evaluate robustness of CL models to train/test augmentations in various cases. We show that different moments allow RobOCLe to capture different properties of deformations, providing higher robustness with no decrease of inference speed.
\end{abstract}

\section{INTRODUCTION}
\label{sec:introduction}

In the last decades, vision models have outperformed human-level accuracy on many recognition benchmarks. 
Deep learning has produced outstanding results by training all  parameters of large models via offline updates over large batches of abundant samples available all at once.

The rise of specialized robots (\eg, indoor domestic and service robots) has driven
the recent advancements in automatic scene analyses.
However, vision models deployed on robotic agents experience new classes and domains specific to users at test time \cite{churamani2020continual}; and classical models cannot be easily personalized without suffering from catastrophic forgetting. %
Hence, Continual Learning (CL) paradigm has emerged to address such adaptation-preservation challenge \cite{lesort2020continual,kirkpatrick2017overcoming}. 

At the same time, robotic agents are equipped with low computation resources and no storage availability due to hardware and privacy constraints \cite{hayes2022online,pellegrini2021continual,knauer2022recall}. 
Therefore, online CL (OCL) methods have gained attention \cite{mai2022} to learn from a non-repeated stream of data and tasks.
In particular, OCL has been recently investigated for low-resource embedded devices with no data storage available \cite{hayes2022online,borghi2023challenges}. However, no extensive study exists at the time of writing. 
Furthermore, when learning new classes, users will only provide a few labelled samples, as the labeling operation is time-consuming and tedious. To tackle this, few-shot (FS) learning \cite{snell2017prototypical} %
should pair with OCL (FS-OCL) \cite{lunayach2022lifelong}.

Finally, after learning new concepts, robotic agents move around the environment (\eg, the user home) and discover the same or other instances of newly learned objects in different conditions. Therefore, vision models should be robust to test-time distortion of objects (\eg, different poses, viewpoints, \etc) and/or environment (\eg, different illumination, room, \etc) \cite{cohen2019certified,zhang2022memo}.
For example, a user may want to update the robotic agent to recognize, \eg, its own pet. The user will likely provide \textit{a few} images of the pet, \eg, lying on the living room carpet; however, the agent should identify the same pet in other poses (\eg, standing in front of a door) and in other domains (\eg, corridor or  kitchen). 

\begin{figure}[tb]
    \centering
    \includegraphics[trim=0cm 9cm 9.25cm 0cm, clip, width=1\linewidth]{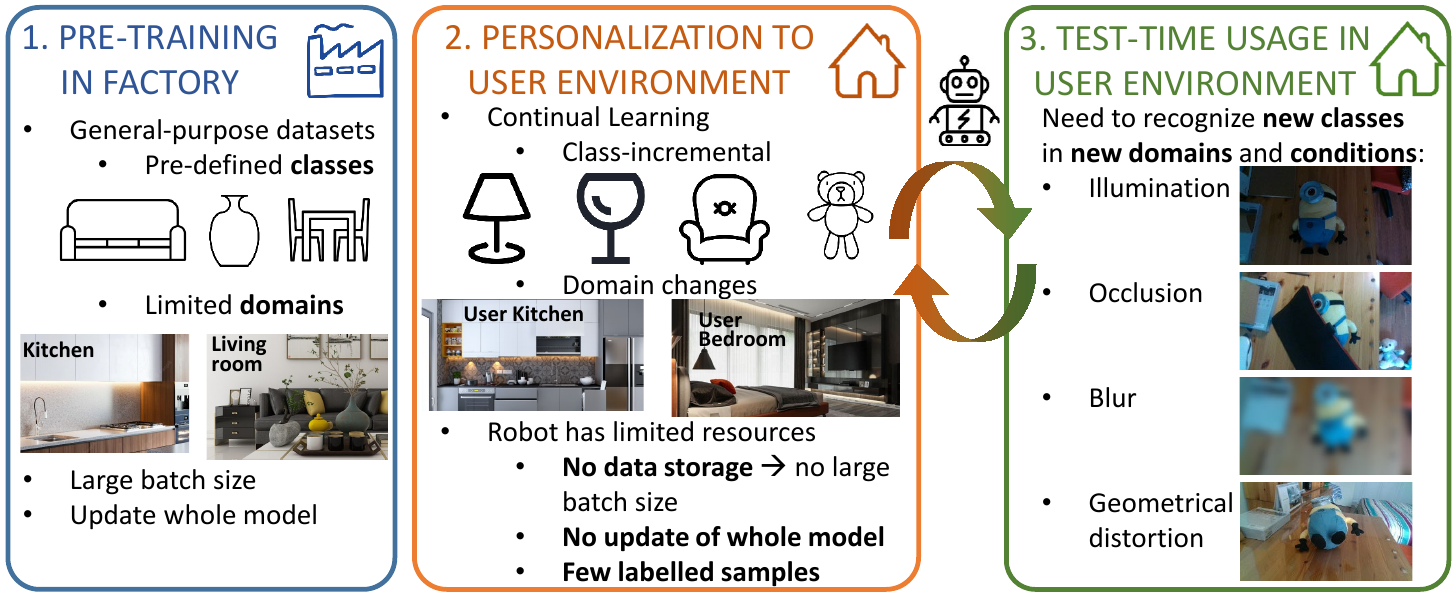}
    \vspace{-15pt}
    \caption{Overview of the considered use-case in 3 stages. 1) A vision recognition model is pre-trained on factory servers on large-scale general-purpose datasets on a pre-defined set of classes and a limited set of acquisition domains. Model pre-training requires large computational power to update the whole model via large batches. 2) The model is embedded into \eg, home robots, and shipped to users. Users customize the recognition model to discover novel classes in their target domain/environment (\eg, user home). Robots do not have large storage due to privacy and storage constraints, and cannot update the whole model due to computational limitations. Large batches are also unavailable, since no storage and low RAM is assumed. Users do not label many samples as it is a tedious operation. 3) The robot navigates the new environment and discovers recently-added classes in unseen domains and conditions, \eg, under variable illumination, occlusion, blurring and geometric distortions. 
    Stages 2) and 3) can alternate at any time.}
     \vspace{-15pt}
\label{fig:use_case}
\end{figure}

In this paper, we introduce the novel scenario of parameter-efficient FS-OCL for low-resource robotic agents for robust test-time performance in variable conditions. \rev{We extend previous setups \cite{hayes2022online} to analyse and improve the robustness of the algorithms to test-time distortions without updating model weights at test time.}
To our knowledge, this challenging and very practical scenario has not been addressed in the existing literature, and in general, little exploration has been carried out on robustness of CL models to variable test-time conditions \cite{wang2022continual,toldo2022learning}. 
The main details of our scenario are depicted in Fig.~\ref{fig:use_case}.

Our contributions are summarized as follows:
\begin{itemize}[leftmargin=*]
    \item We tackle a new FS-OCL paradigm for low-resource robots where storage is unavailable and computation power is low (\ie, most of model weights cannot be updated).
    \item We introduce a new parameter-efficient FS-OCL method (RobOCLe) suitable for mobile robotic agents and FS data. RobOCLe extracts high order statistics from embeddings, building more reliable feature representations robust to variable domain and object conditions. RobOCLe shows consistent improvements against 10 OCL baselines on 4 benchmarks and 16 backbones. RobOCLe shows a room-aware relative accuracy gain (RARG, Sec.~\ref{sec:setup}) of average 65.8\% on same-domain data and 16.4\% on other-domain data compared to the best \rev{competing approach}, while decreasing inference FPS by less than 0.5\%. 
    \item We present a new evaluation paradigm to determine robustness of vision models, and particularly CL models, over a suite of real and synthetic augmentations, resembling practical use cases. RobOCLe shows average RARG of 18.3\% on controlled augmentations of other-domain data.
\end{itemize}

\section{RELATED WORK}
\label{sec:related}

\textbf{Continual Learning} has many definitions.
Classical CL assumes data released at incremental steps and  multiple training epochs performed with large batch sizes updating the whole net \cite{kirkpatrick2017overcoming,zenke2017continual,de2021continual,lopez2017gradient}.
Online CL (OCL) assumes, instead, that data comes as a non-repeated stream of small-batch samples \cite{mai2022,hayes2019memory,mai2021supervised,hayes2020lifelong,wolfe2022cold,pellegrini2020latent}. OCL methods are mainly based on replay due to the inherent difficulty to prevent forgetting.
Recently, OCL for embedded devices \cite{hayes2022online,borghi2023challenges,ravaglia2021tinyml,demosthenous2021continual} further restricts the scope to parameter-efficient updates with unitary batch size due to low-resource devices.
In parallel, few-shot CL (FS-CL) \cite{ayub2020cognitively,ayub2021f,ayub2020tell,mazumder2021few,von2021learning} and FS-OCL \cite{lunayach2022lifelong} train a model to recognize new classes based on few labelled data only, via fast and efficient model updates with minimal human effort.

Along all these definitions, three main scenarios have been considered \cite{vandeven2022three}: 1) class-incremental (CI) where new classes are introduced to models over time; 2) domain-incremental (DI) where the same problem is learned in different contexts (\eg, with a fixed set of classes); and 3) task-incremental (TI), where new distinct tasks are presented to models.

In our paper, we focus on the most practical and less-explored FS-OCL for low-resource devices and few-shot data in the CI setup, with domain changes happening at both fine-tuning and testing stages.
OCL, FSCL and FSOCL methods generally assume replay data \cite{chaudhry2018riemannian,shim2021online,chaudhry2019efficient,mai2022}, violating the important assumption that no storage is guaranteed. Therefore, we rely on the baselines introduced in \cite{hayes2022online}.
Additionally, we evaluate robustness of agents at test time.

\textbf{Test-Time Performance} is heavily influenced by the domain gap with respect to the training set \cite{wang2018deep,toldo2020unsupervised}. Test-Time Augmentation (TTA) approaches use data augmentation at test time to obtain greater robustness
\cite{cohen2019certified,zhang2022memo},
improved accuracy \cite{shanmugam2021better,krizhevsky2017imagenet,he2016deep,lu2022improved,ashukha2021mean,chun2022cyclic}, %
or estimates of uncertainty \cite{wang2019aleatoric,conde2022adaptive}. %
TTA pools predictions from several transformed versions of a given test input to obtain a \textit{smoothed} prediction.
In particular, %
\cite{krizhevsky2017imagenet,he2016deep,lu2022improved} average the predictions obtained over augmented views (\eg, cropped, rotated, \etc) of input samples. %
\cite{ashukha2021mean} averages embeddings obtained over augmented views of the input samples.
\cite{shanmugam2021better,chun2022cyclic} propose a learnable aggregation of TTAs.
\cite{kim2020learning,chun2022cyclic} learn a selection policy for TTAs. %
\cite{enomoto2022augnet} enhances images to make them recognition-friendly.

Test-Time Training (TTT) is a special case of Unsupervised Domain Adaptation (UDA) where a model trained on a source domain is adapted to an unlabeled target domain without accessing source data \cite{goyal2022test,mirza2022actmad,su2022revisiting}. %
Many approaches design novel loss functions to train the model and reduce domain shift, requiring access to the full target set.
Recently, \cite{zhang2022memo} performs TTT on single images 
coupling TTA with TTT to improve test-time robustness.
Continual TTT \cite{wang2022continual,toldo2022learning} has been recently considered.

Our work is linked to this, with some key differences:
\begin{itemize}[leftmargin=*]
    \item We do not assume the test set to be available all at once, as in practical applications where an agent encounters new samples and domains over time.  
    \item Our scope is different. TTA boosts model performance at test time aggregating multiple predictions. TTT updated the model at test time. Our aim is to propose an OCL method with increased robustness over variable test sets.
    \item We do not compute multiple TTAs of the same input sample, but rather mimic a deployment scenario where robots experience variable domains with no clear delimitation.
    \item We extract rich information from inputs, to reduce the need for TTA/TTT and maintain same inference time (we use one inference step per sample, opposed to TTA/TTT).
\end{itemize}

\noindent\textbf{Pooling} has been one of the key enablers for deep learning models mapping high dimensional features into a summary of active features \cite{sum1}. %
Multiple schemes have been proposed
to extract the most useful information from the input data.
We compare our pooling mechanism with the following schemes:
average pooling (AVG) \cite{lecun1998gradient};
max pooling (MAX) \cite{riesenhuber1999hierarchical};
a linear combination of AVG and MAX (MIX) \cite{zhou2021mixed};
a dropout-inspired probabilistic selection of activations based on 
a multinomial distribution fit (STOCHASTIC) \cite{zeiler2013stochastic}; 
a concatenation of top-k\% of features to discover information spread over the spatial map (RAP) \cite{bera2020effect};
a concatenation of AVG and MAX pooling (AVGMAX) \cite{monteiro2020performance};
an average of the $p$-th order non-central moment ({L$_{\mathbf{p}}$-norm}) \cite{feng2011geometric};
and iSQRT-COV, which computes second-order covariance pooling via iterative  matrix square root normalization \cite{li2018towards}.
In our work, we consider high order central moments \rev{\cite{michieli2023online}} to discover robust features from limited input data.

\section{METHODS}
\label{sec:method}

\rev{We investigate OCL from sequential data streams. We evaluate existing approaches and propose a new method (RobOCLe) to overcome some of the challenges that CL methods face when tested in novel conditions.}

\rev{In our setup, a model learns from streamed examples, seen one at a time with no repetitions, since many applications cannot store samples due to the continuous nature of the data stream and the limited time and storage to process it. Over time, the learner sees new classes but no task identifier is provided, since the agent should infer the class regardless of the specific task information.}
\rev{On resource-constrained devices, CL methods should be both data and computation efficient: they should learn from few-shot labelled data and adhere to strict memory and time constraints. Finally, agents will operate in uncontrolled environments, therefore, CL methods should be robust to (severe) test-time corruptions.}

\subsection{RobOCLe: \textbf{Rob}ust \textbf{O}nline \textbf{C}ontinual \textbf{L}earning}

To tackle our scenario, we propose a new \textbf{Rob}ust \textbf{O}nline \textbf{C}ontinual \textbf{Le}arning method (RobOCLe), comprising of 3 main blocks: a feature extractor $G(\cdot)$, a pooling scheme $P(\cdot)$ and a classifier $F(\cdot)$ to categorize samples from their extracted features.
We incrementally train a neural network $F(P(G(\mathbf{x}_{n,k})))$ via supervised OCL updates over subsequent tasks to generate predictions $\hat{y}_{n,k}$, where $\mathbf{x}_{n,k}$ is the $n$-{th} input sample, ${n\in[N_k]}$, of the $k$-{th} task, {$k \in [K]$}. 

Following traditional transfer learning setups, $G(\cdot)$ is a backbone network (\eg, ResNets \cite{he2016deep}) pre-trained on a server on public benchmarks, with large batch sizes over many epochs updating the whole model (point 1 of Fig.~\ref{fig:use_case}).
When $G(\cdot)$ is embedded in robotic agents, it experiences new domains (\eg, user environments) compared to training ones.

Additionally, users want to personalize models to recognize personal objects. Therefore, a novel classifier $F(\cdot)$ adapts the general features extracted by $G(\cdot)$ to the class set of particular users (point 2 of Fig.~\ref{fig:use_case}).
Users show few labeled samples of the new classes to recognize, and robots do not have storage or large computation resources. Therefore, $F(\cdot)$ is trained using online updates, seeing a non-repeated stream of few-shot labelled samples using a frozen feature extractor $G(\cdot)$. 
In the experiments, we consider the most challenging case of a single class being learned per task.

\begin{figure*}[t]
    \centering
    \setlength{\tabcolsep}{5pt}
    \renewcommand{\arraystretch}{1}
\begin{tabular}{cccc}
    & R=1 (Average AVG \cite{lecun1998gradient}) &  R=2 (Variance) &  R=3 (Skewness) \\
    {\begin{sideways}\quad \ \scalebox{1}{Frequency}\end{sideways}} &
     \includegraphics[width=0.25\linewidth, trim=1.5cm 0.18cm 1.6cm 0.9cm, clip]{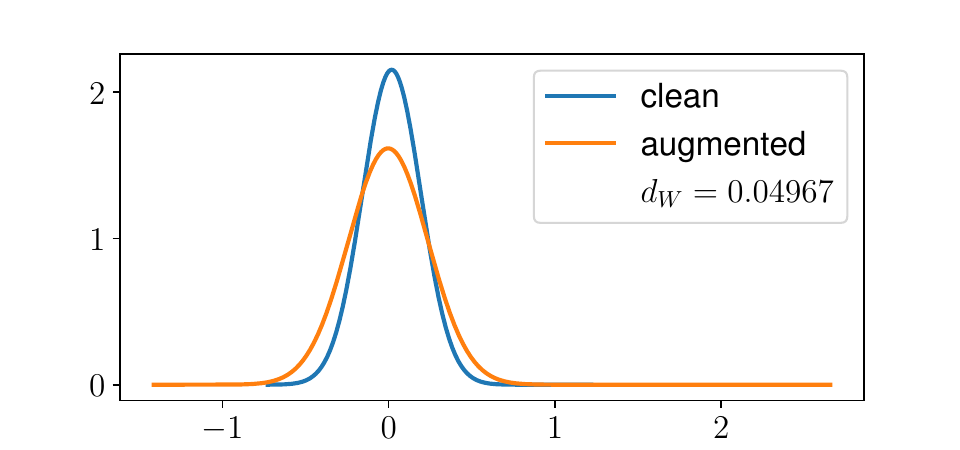} &
     \includegraphics[width=0.25\linewidth, trim=1.5cm 0.18cm 1.6cm 0.9cm, clip]{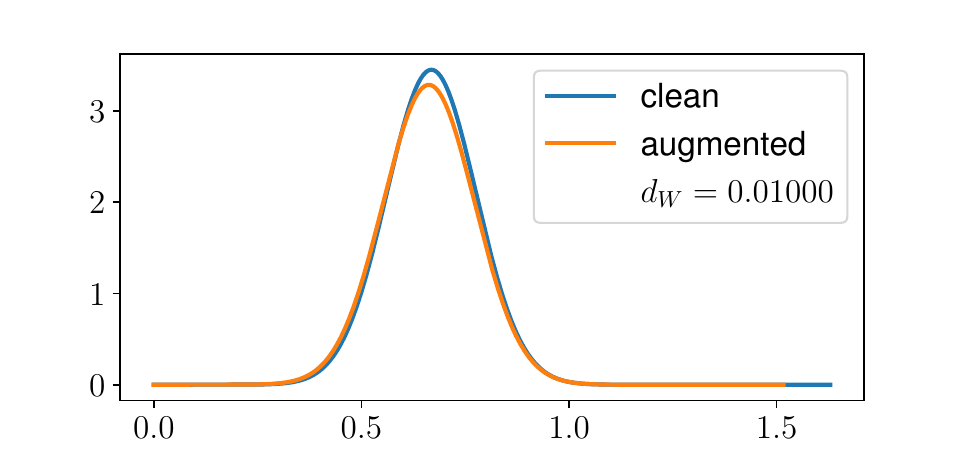} &
     \includegraphics[width=0.25\linewidth, trim=1.5cm 0.18cm 1.6cm 0.9cm, clip]{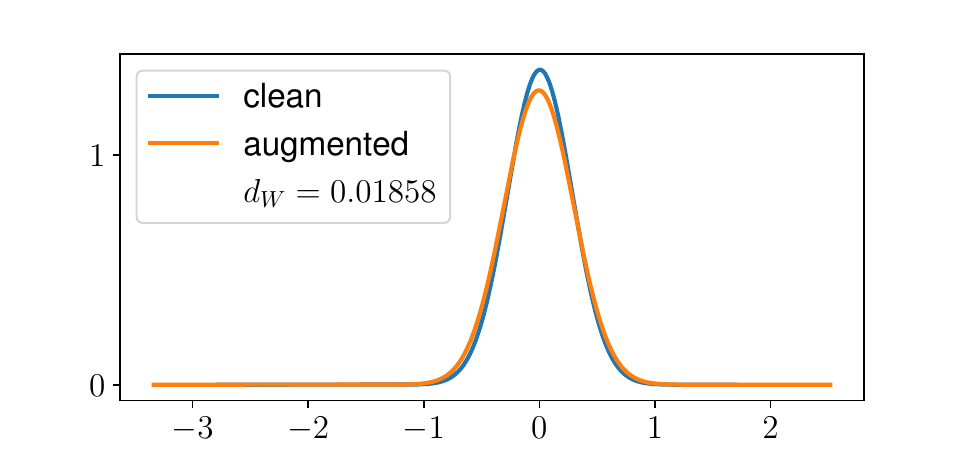} \\
     & \scalebox{1}{Value} & \scalebox{1}{Value} & \scalebox{1}{Value} \\
\end{tabular}
\vspace{-4pt}
    \caption{Distribution of statistical moments in $\mathbf{z}_n$. $d_W$: Wasserstein distance.}
     \vspace{-10pt}
    \label{fig:moments}
\end{figure*}

\textbf{High order pooling.} In order to apply $F(\cdot)$ over the pre-trained backbone $G(\cdot)$, a pooling scheme is generally inserted to summarize the input information and reduce the feature map size.
At test time, the robotic agent should recognize  personal classes in a variety of conditions (\eg, illumination, occlusion, blur, pose). Therefore, the overall OCL procedure should be robust to such factors of variations at test time.
With this aim in mind, we perform the pooling $P(\cdot)$, employing higher order statistical moments to increase the amount of clues extracted from the input samples. In particular, $P(\cdot)$ computes and concatenates the first $R$ statistical moments from the output of $G(\cdot)$. 
Such moments characterize the distribution $\mathcal{G}$ of $g \triangleq G(\mathbf{x}_{n,k}) \in \mathbb{R}^{h\times w\times d}$ where $h$ and $w$ are the spatial sizes of features and $d$ is the number of channels, and thus capture rich spatial statistics to improve  recognition accuracy. More formally, we employ

\begin{equation}
    P\left(g\right) \! {=} \!  \concat   \left(  \mu, 
    {{E_\mathcal{G} \left[  (g {-}\mu)^2  \right]^\frac{1}{2}}}\!,
    \concat_{r=3}^R \! E_\mathcal{G} \left[
    \frac{g{-}\mu}
    {E_\mathcal{G} \left[  (g {-}\mu)^2  \right]^\frac{1}{2}}
    \right]^{\!r}
    \right)\!,
\end{equation}
where $E_\mathcal{G}[\cdot]$ is the expectation over $\mathcal{G}$, $\mu$ is the empirical mean,
and $\concat(\cdot)$ denotes the concatenation operation. That is, $P(g) \in \mathbb{R}^{R \cdot d}$ concatenates the first $R$ moments of $g$.
In our case, we feed ${R=3}$ moments to the  classifier.
Fig.~\ref{fig:moments} shows the distribution of the first statistical moments of features extracted from clean vs.\ augmented samples. 
AVG has more variability (higher Wasserstein distance) than higher moments when changing domain.  In Sec.~\ref{sec:results} we confirm that high order moments increase robustness of OCL models and their invariance to augmentations on both same-domain and other-domain test data (\ie, when models receive test data in other conditions than at train time).

\textbf{Similarity estimation.} 
The classifier $F(\cdot)$ should be representative of the new classes and extremely lightweight for on-device deployment. We extend the Nearest Classifier Mean (NCM) and  the Linear Discriminant Analysis (LDA) 
to support the streaming setup \cite{hayes2020lifelong} and the high order pooling. Streaming NCM and LDA are denoted by NCM and SLDA. 

RobOCLe$_\mathrm{NCM}$ computes a running mean feature vector per class (\ie, the $c$-th class prototype $\mathbf{m}_c \in \mathbb{R}^{R\cdot d}$, ${\forall c\in\mathcal{C}}$) each with an associated counter denoting the number of samples employed to compute each average value ($\mathbf{t}_c$). Given a new data vector $\mathbf{x}_{n,k}$ with the associated label $y_{n,k}$, we embed it to $\mathbf{z}_{n,k} \triangleq P(G(\mathbf{x}_{n,k})) \in \mathbb{R}^{R\cdot d}$, and  update the class mean and associated counter by
\begin{equation}
    \mathbf{m}_{y_{n,k}} \leftarrow \frac{\mathbf{t}_{y_{n,k}} \cdot \mathbf{m}_{y_{n,k}} + \mathbf{z}_{n,k} }{\mathbf{t}_{y_{n,k}} + 1},
    \quad
    \mathbf{t}_{y_{n,k}} \leftarrow \mathbf{t}_{y_{n,k}} + 1.
\end{equation}
For inference on new samples, RobOCLe$_\mathrm{NCM}$ assigns the label of the nearest prototype according to its $\ell_2$ distance on the enriched feature space. The original NCM has shown to be a simple yet effective baseline in CL \cite{rebuffi2017icarl,mai2021supervised,hayes2022online,mensink2013distance}.

RobOCLe$_\mathrm{SLDA}$ computes one channel-wise covariance matrix of features shared across classes ($\mathbf{\Sigma} \in \mathbb{R}^{d\times d}$) that is updated online via \cite{dasgupta2007line}. During inference, SLDA assigns to a new sample the label of the closest Gaussian model in the feature space defined using the running class means and shared $\mathbf{\Sigma}$. RobOCLe$_\mathrm{NCM}$ is a special case of RobOCLe$_\mathrm{SLDA}$ where $\mathbf{\Sigma}$ is equal to the identity matrix.
We use the implementation from \cite{hayes2020lifelong} to update the  $\mathbf{\Sigma}$ and compute predictions. 
RobOCLe$_\mathrm{SLDA}$ runs inference on a new sample\footnote{To simplify notation, we remove task index $k$.} $\mathbf{x}_n$ by ${F(P(G(\mathbf{x}_n))) = \mathbf{W} \mathbf{z}_n + \mathbf{b}}$, where $\mathbf{W}\in\mathbb{R}^{|\mathcal{C}|\times R \cdot d}$ and ${\mathbf{b} \in \mathbb{R}^{|\mathcal{C}|}}$, where $|\cdot|$ is the set cardinality.
Rows of $\mathbf{W}$ are computed by $\mathbf{w}_c = \mathbf{\Lambda} \mathbf{m}_c^T$, and elements of $\mathbf{b}$ are computed by ${b_c = -0.5 (\mathbf{m}_c \cdot \mathbf{\Lambda}\mathbf{m}_c) = (\mathbf{m}_c \cdot \mathbf{w}_c)}$ %
with the shrinking approximation $\mathbf{\Lambda} = [ (1-\epsilon) \mathbf{\Sigma} + \epsilon \mathbf{I} ]^{-1}$ with parameter $\epsilon = 10^{-4}$. Running covariance \cite{dasgupta2007line} is then computed by 
\begin{equation}
    \mathbf{\Sigma}_{n+1} = \frac{n\mathbf{\Sigma}_n+\delta_n}{n+1},
\ \ 
    \delta_n = \frac{n(\mathbf{z}_n-\mathbf{m}_{y_n})(\mathbf{z}_n-\mathbf{m}_{y_n})^T}{n+1}. 
\end{equation}

\subsection{OCL Baselines}

We describe  several OCL methods used to update $F(\cdot)$, mostly focusing on data-free approaches.
\noindent\textbf{Fine-Tuning (FT)} updates a fully-connected output layer using SGD and CE loss with no mechanisms to prevent forgetting \cite{hayes2022online}.
\noindent\textbf{Online Perceptron (PRCPT)} keeps one weight vector for each class \cite{hayes2022online}, initialized to the first sample of the class. After that, when the model misclassifies a new sample, the class weight vector and the weight vector of the misclassified class with the highest score are updated. During inference, a label is assigned by taking the $\argmax$ over the dot product between weights and input vector.
\noindent\textbf{Online Centroid-Based Concept Learning (CBCL)} extends NCM with multiple class prototypes \cite{ayub2020cognitively}. At inference, CBCL searches for the weighted nearest neighbor, where class weights are inversely proportional to the number of samples seen so far for that class.
We use the default parameters suggested in \cite{ayub2021f}, \ie, a distance threshold of $17$, a nearest neighbor value of $1$, and a maximum buffer size of $44$ prototypes. We noticed no clear difference compared to NCM, as in \cite{hayes2022online}.
\noindent\textbf{Streaming One-vs-Rest (SOvR)} measures how close a new input is to a class mean vector while also considering its distance to examples from other classes \cite{hayes2022online}, which is reminiscent of SVMs. 
\noindent\textbf{SQDA} extends SLDA estimating one covariance matrix for each class (\ie, Quadratic). The drawbacks are increased memory consumption and need for many samples per class to estimate reliable covariance matrices \cite{anagnostopoulos2012online}.
\noindent\textbf{Streaming Gaussian Na\"ive Bayes (SNB)} estimates a running variance vector per class \cite{welford1962note} (\ie, diagonal covariance matrices assuming independent features). It requires significantly less memory than SQDA.
\noindent\textbf{Online iCaRL}  maintains a class-balanced memory buffer 
\cite{hayes2020lifelong,rebuffi2017icarl} %
and randomly replaces an example from the most represented class with a new sample when the buffer is full. During training, it randomly draws examples from the buffer, combines them with the new sample and makes a single update. Similar to other methods, we train a  linear output layer. While effective, it requires storing sensitive samples on devices. We use a buffer size of 1K samples (iCarl) or of 2 samples per class (iCarl-2pc).

\section{Experimental Setup}
\label{sec:setup}

\begin{figure}[tb]
    \centering
    \includegraphics[trim=0cm 11.2cm 6.6cm 0cm, clip, width=1\linewidth]{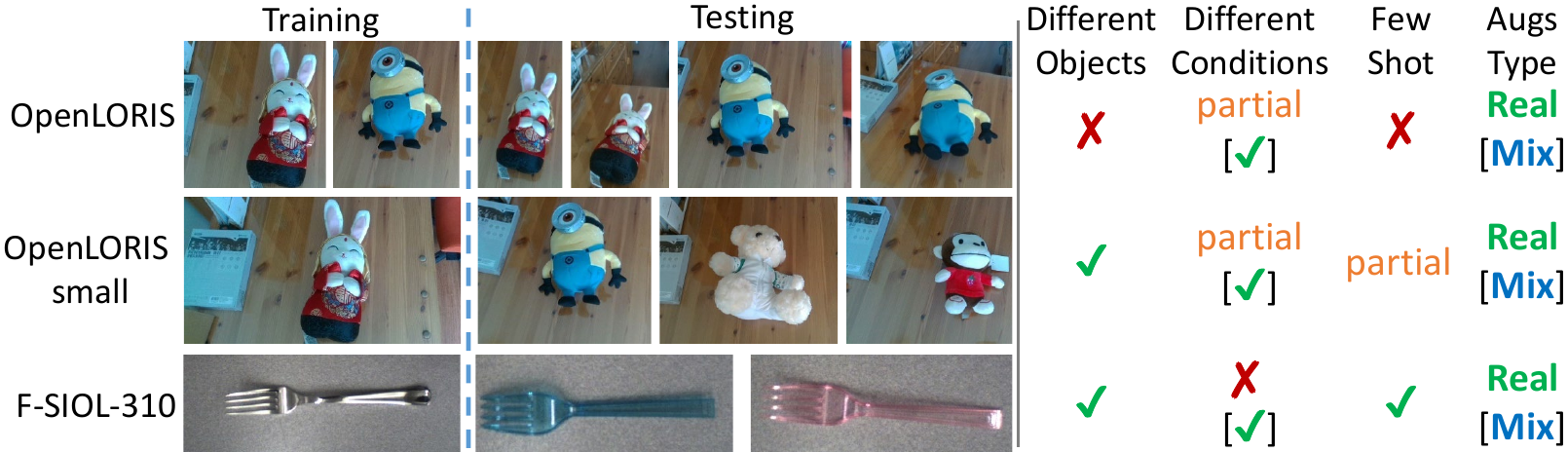}
    \vspace{-15pt}
    \caption{Datasets employed in our work. Right-hand side table summarizes differences between train and test sets. [Text within squared brackets] specifies how properties change when our augmentations are applied (Sec.~\ref{sec:setup}).}
    \vspace{-17pt}
\label{fig:datasets}
\end{figure}

\noindent\textbf{Datasets.} We analyse our novel setup on F-SIOL-310 \cite{ayub2021f} and OpenLORIS(-Objects) \cite{she2020openloris}, which are robot-acquired datasets specifically designed for CL and FS, and their  train and test sets have different properties. A visual summary is reported in Fig.~\ref{fig:datasets}.
We consider a class-IID setup, where all samples are ordered by class and shuffled within each class.

\textit{OpenLORIS} collects 121 instances including 40 categories of daily necessities objects under 20 scenes from variable camera-object distance/angle. 
The dataset comprises four environmental factors: \textit{illumination} variability during recording, \textit{occlusion} percentage of the objects, object pixel size in each frame (\ie, \textit{scale} of objects), and \textit{clutter} of the scene. Each factor has three levels of intensity, yielding around 550,000 total RGB samples.
Following \cite{hayes2019memory,hayes2022online}, we consider two splits: instance and instance-small. 

The instance split (referred to as OpenLORIS) presents videos (\ie, image frames from the same video) of shuffled object instances to learners one at a time. For example, a learner experiences a video of toy\#1, then a video of hat\#3, then a video of toy\#2, and so on. The learner experiences temporally-correlated data with repeating classes. This split uses the original train-val division provided by the authors. 

The instance-small split (OpenLORIS-small) evaluates low-shot out-of-domain generalization ability of models. When learning a new class, frames from a single video of an object instance are shown to the learner. When testing, the model is evaluated on all test data containing the classes seen so far. Test data contains same object instance seen under different domains (\ie, factors) as well as other object instances from the same classes seen under different domains. Challenges of this setup are: low-shot labelled learning from correlated data, generalization to different object instances of same class, generalization to unseen domains.
This setup is often encountered in practice. For example, given a new pet, users provide a few labels and a classifier should recognize the pet in different domains and conditions.

\textit{F-SIOL-310} is specifically designed for Few-Shot Incremental Object Learning. We employ this dataset to further evaluate the ability of models to generalize from extremely small labelled data. It contains 310 instances of 22 household categories with no object instances overlapping between train and test sets.
We consider two splits, respectively with 5 and 10 shots (\ie, 5 or 10 labelled training samples).

\noindent\textbf{Objects, Conditions \& Augmentations.} The considered datasets span a wide range of setups with variable factors.
OpenLORIS uses the whole dataset and the same objects are shown in train and test sets. Images are acquired under some different conditions (illumination, occlusion, scale, clutter). However, co-existence of conditions is limited.
OpenLORIS-small restricts the set of samples and has only one instance of objects in training.
F-SIOL-310 uses 5 or 10 training samples, and has different objects between train and test sets. However, objects are acquired with similar conditions (same background, viewpoint/scale, no occlusion, no clutter).

\begin{figure}[tb]
    \centering
    \includegraphics[trim=0cm 6.75cm 16.2cm 8.9cm, clip, width=1\linewidth]{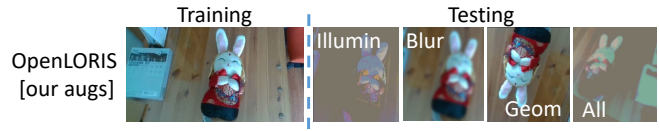}
    \vspace{-20pt}
    \caption{Synthetic data augmentations employed to analyse different conditions in which objects could be found.}
    \vspace{-17pt}
\label{fig:augmentations}
\end{figure}

To cope with the limited variability of existing datasets, we introduce some augmentations at either train or test splits.
A visual summary is shown in Fig.~\ref{fig:augmentations}: we mimic several conditions as described next.%
1) Illumination changes via color jittering with parameters chosen uniformly in the following ranges: brightness in $[0.5,1.5]$, contrast in $[0.5,1]$, saturation in $[0.5, 1.5]$, and hue in $[-0.1,0.1]$.   %
2) Image blur (\eg, due to dust on camera or wrong focus) via Gaussian noise with $11\times11$ kernel and standard deviation (std) in $[0.1,0.5]$.
3) Geometrical changes via a combination of 
a) affine transformation keeping image center invariant, with degree in $[-30,30]$, translate shift in $[-0.2,0.2]$ for both horizontal and vertical axes, scale in $[0.8,1.2]$, and shear in $[-0.1,0.1]$);
b) random perspective with probability $0.2$ and distortion amount $0.2$, 
c-d) horizontal and vertical flip with probability $0.5$ and $0.3$, respectively.
4) Last, we also consider all of the above together.
To ensure fair comparisons among OCL methods, all samples are augmented with the same parameters in each set of experiments.

All the results are averaged across $5$ class orderings, although std is not reported to improve readability.
Offline training upper bounds are not considered, because the classifier $F(\cdot)$ is different for each approach.

\noindent\textbf{Metrics.}
We evaluate both plasticity and forgetting in comparison of OCL methods \cite{mai2022}. All metrics are computed on test sets at the end of  training phases.
We compute accuracy (Acc \%, $\uparrow$) to assess the final model performance across all classes. \um{\um{Often, we report the room-aware relative gain, RARG, ($\Delta_R$, \%)  of  model$_2$ with accuracy $\mathrm{Acc}_2$ compared to a model$_1$ with $\mathrm{Acc}_1$ with respect to the available room of improvement, defined by ${\Delta_R\triangleq(\mathrm{Acc}_2- \mathrm{Acc}_1) / (100 - \mathrm{Acc}_1)}$.}}
In addition, we also measure: 
(1) backward transfer (BwT, $\uparrow$), which tracks the influence that learning a new task has on the preceding tasks’ performance, to measure stability; 
(2) forgetting (Forg, $\downarrow$), which averages  difference of class-wise Acc at the last step and the best previous class-wise Acc;
(3) plasticity (Pla, $\uparrow$) as the average Acc achieved on each task evaluated right after learning that task; 
(4) training time (TTime [min]); 
and (5)  frames per second (FPS) in inference. 
Evaluation is carried out using an  NVIDIA GeForce RTX 2080Ti GPU supported by an Intel(R) Xeon(R) Gold 5218R CPU@2.10GHz with 80 cores and 252GB RAM. Nonetheless, relative considerations remain the same on hardware mounted on robotic agents. \rev{All results are averaged across 5 runs with different seeds: only the mean value is shown since the standard deviation is negligible ($\leq0.16$ in all cases).}

\section{EXPERIMENTAL VALIDATION}
\label{sec:results}

\begin{table*}[t]
\setlength{\tabcolsep}{1.1pt}
  \centering
  \caption{Acc on same-domain OpenLORIS data on $16$ backbones and $10$ OCL baselines. MN: MobileNet, EN: EfficientNet, RN:ResNet.} %
  \vspace{-8pt}
  \resizebox{0.93\linewidth}{!}{%
  \begin{tabular}{lcccccccccccccccca}
  \toprule
          & \textbf{MN-S} & \textbf{MN-L} & \textbf{EN-B0} & \textbf{EN-B1} & \textbf{RN18} & \textbf{RN34} & \textbf{RN50} & \textbf{RN101} & \textbf{RN152} & \textbf{Swin-T} & \textbf{Swin-S} & \textbf{Swin-B} & \textbf{ViT-B16} & \textbf{ViT-B32} & \textbf{ViT-L16} & \textbf{ViT-L32} & \textbf{Avg} \\
    \midrule
    FT & 84.22 & 93.38 & 96.14 & 97.89 & 83.35 & 83.84 & 94.19 & 95.43 & 94.33 & 95.68 & 96.07 & 96.04 & 59.07 & 67.27 & 92.47 & 62.85 & 87.01 \\
    PRCPT \cite{hayes2022online} & 74.49 & 89.75 & 94.02 & 95.94 & 80.47 & 78.58 & 92.18 & 91.24 & 92.35 & 93.40 & 94.21 & 94.46 & 48.70 & 58.51 & 85.62 & 55.19 & 82.44 \\
    SNB \cite{welford1962note} & 31.12 & 37.84 & 77.96 & 83.91 & 1.51  & 0.84  & 0.00  & 0.00  & 0.00  & 87.30 & 86.94 & 86.14 & 3.51  & 4.60  & 42.90 & 5.76  & 34.40 \\
    SOvR \cite{hayes2022online} & 37.42 & 60.17 & 73.92 & 80.65 & 34.65 & 31.77 & 71.54 & 64.57 & 67.68 & 80.03 & 79.01 & 77.92 & 4.91  & 18.83 & 62.61 & 22.02 & 54.23 \\
    NCM/CBCL \cite{mensink2013distance,ayub2020cognitively} & 72.89 & 81.83 & 85.94 & 88.34 & 79.69 & 80.47 & 84.62 & 83.98 & 84.12 & 87.77 & 87.54 & 87.94 & 64.99 & 63.38 & 79.47 & 68.09 & 80.07 \\
    SQDA \cite{anagnostopoulos2012online} & 77.71 & 55.84 & 2.45  & 6.16  & 81.59 & 81.36 & 5.66  & 24.24 & 1.64  & 83.84 & 62.53 & 63.27 & 8.91  & 15.74 & 3.53  & 7.50  & 36.37 \\
    \hdashline
    iCaRL \cite{rebuffi2017icarl} & 91.76 & 95.54 & 97.60 & 98.06 & 92.76 & 93.21 & 97.18 & 97.47 & 97.63 & 97.65 & 97.57 & 97.98 & 86.43 & 89.61 & 95.52 & 89.61 & 94.72 \\
    iCaRL (2pc) \cite{rebuffi2017icarl} & 89.29 & 95.09 & 97.07 & 96.43 & 91.21 & 92.34 & 96.50 & 96.99 & 96.79 & 97.13 & 97.13 & 97.68 & 79.41 & 82.49 & 93.34 & 81.78 & 92.54 \\
    \midrule

    SLDA \cite{hayes2020lifelong} & 95.57 & 97.93 & 98.83 & 98.98 & 95.01 & 95.47 & 99.00 & 99.10 & 99.13 & 98.25 & 98.18 & 98.85 & 96.74 & 96.20 & 98.69 & 97.04 & 97.69 \\
    RobOCLe$_{\mathrm{SLDA}}$ (ours) & \textbf{98.20} & \textbf{99.37} & \textbf{99.70} & \textbf{99.72} & \textbf{97.65} & \textbf{97.96} & \textbf{99.69} & \textbf{99.78} & \textbf{99.78} & \textbf{99.28} & \textbf{99.31} & \textbf{99.65} & \textbf{99.16} & \textbf{99.02} & \textbf{99.73} & \textbf{99.33} & \textbf{99.21} \\
    ($\Delta_R$ [\%]) & \scriptsize\color{OliveGreen}(+59.5)  & \scriptsize\color{OliveGreen}(+69.4)  & \scriptsize\color{OliveGreen}(+74.8)  & \scriptsize\color{OliveGreen}(+72.3)  & \scriptsize\color{OliveGreen}(+52.8)  & \scriptsize\color{OliveGreen}(+54.9)  & \scriptsize\color{OliveGreen}(+69.1)  & \scriptsize\color{OliveGreen}(+75.6)  & \scriptsize\color{OliveGreen}(+75.2)  & \scriptsize\color{OliveGreen}(+58.6)  & \scriptsize\color{OliveGreen}(+62.0)  & \scriptsize\color{OliveGreen}(+69.9)  & \scriptsize\color{OliveGreen}(+74.2)  & \scriptsize\color{OliveGreen}(+74.3)  & \scriptsize\color{OliveGreen}(+79.6)  & \scriptsize\color{OliveGreen}(+77.4)  & \scriptsize\color{OliveGreen}(+65.8) \\
    \bottomrule
    \end{tabular}%
  \label{tab:indomain}%
  }
   \vspace{-5pt}
\end{table*}%

\begin{table*}[t]
\setlength{\tabcolsep}{0.9pt}
  \centering
  \caption{Accuracy on real other-domain few-shot datasets on ResNets and ViTs backbones.}
   \vspace{-8pt}
  \resizebox{\linewidth}{!}{%
    \begin{tabular}{lcccccccccccccccccccccccc}
    \toprule
          & \multicolumn{8}{c}{\textbf{OpenLORIS-small}}                           & \multicolumn{8}{c}{\cellcolor[rgb]{ .851,  .851,  .851}\textbf{F-SIOL-310 (5-shots)}} & \multicolumn{8}{c}{\textbf{F-SIOL-310 (10-shots)}} \\
          & \scriptsize\textbf{RN50} & \scriptsize\textbf{RN101} & \scriptsize\textbf{RN152} & \scriptsize\textbf{ViT-B16} & \scriptsize\textbf{ViT-B32} & \scriptsize\textbf{ViT-L16} & \scriptsize\textbf{ViT-L32} & \cellcolor{LightBlue}\scriptsize\textbf{Avg} & \cellcolor[rgb]{ .851,  .851,  .851}\scriptsize\textbf{RN50} & \cellcolor[rgb]{ .851,  .851,  .851}\scriptsize\textbf{RN101} & \cellcolor[rgb]{ .851,  .851,  .851}\scriptsize\textbf{RN152} & \cellcolor[rgb]{ .851,  .851,  .851}\scriptsize\textbf{ViT-B16} & \cellcolor[rgb]{ .851,  .851,  .851}\scriptsize\textbf{ViT-B32} & \cellcolor[rgb]{ .851,  .851,  .851}\scriptsize\textbf{ViT-L16} & \cellcolor[rgb]{ .851,  .851,  .851}\scriptsize\textbf{ViT-L32} & \cellcolor{LightBlue}\scriptsize\textbf{Avg} & \scriptsize\textbf{RN50} & \scriptsize\textbf{RN101} & \scriptsize\textbf{RN152} & \scriptsize\textbf{ViT-B16} & \scriptsize\textbf{ViT-B32} & \scriptsize\textbf{ViT-L16} & \scriptsize\textbf{ViT-L32} & \cellcolor{LightBlue}\scriptsize\textbf{Avg} \\
    \midrule
    FT & 13.64 & 14.81 & 16.05 & 2.08  & 2.08  & 2.51  & 2.08  & \cellcolor{LightBlue}7.61  & \cellcolor[rgb]{ .851,  .851,  .851}40.72 & \cellcolor[rgb]{ .851,  .851,  .851}27.65 & \cellcolor[rgb]{ .851,  .851,  .851}31.96 & \cellcolor[rgb]{ .851,  .851,  .851}8.56 & \cellcolor[rgb]{ .851,  .851,  .851}9.87 & \cellcolor[rgb]{ .851,  .851,  .851}19.67 & \cellcolor[rgb]{ .851,  .851,  .851}8.63 & \cellcolor[rgb]{ .851,  .851,  .851}\cellcolor{LightBlue}21.01 & 30.25 & 26.75 & 32.83 & 5.00  & 7.33  & 14.42 & 7.25  & \cellcolor{LightBlue}17.69 \\
    PRCPT \cite{hayes2022online} & 27.45 & 21.68 & 21.36 & 2.08  & 2.18  & 6.64  & 2.08  & \cellcolor{LightBlue}11.92 & \cellcolor[rgb]{ .851,  .851,  .851}32.16 & \cellcolor[rgb]{ .851,  .851,  .851}24.12 & \cellcolor[rgb]{ .851,  .851,  .851}34.31 & \cellcolor[rgb]{ .851,  .851,  .851}4.84 & \cellcolor[rgb]{ .851,  .851,  .851}5.42 & \cellcolor[rgb]{ .851,  .851,  .851}7.39 & \cellcolor[rgb]{ .851,  .851,  .851}4.84 & \cellcolor[rgb]{ .851,  .851,  .851}\cellcolor{LightBlue}16.15 & 38.25 & 24.42 & 32.75 & 5.17  & 5.17  & 9.75  & 4.92  & \cellcolor{LightBlue}17.20 \\
    SNB \cite{welford1962note} & 0.57  & 0.58  & 0.25  & 3.10  & 4.19  & 23.52 & 5.16  & \cellcolor{LightBlue}5.34  & \cellcolor[rgb]{ .851,  .851,  .851}4.90 & \cellcolor[rgb]{ .851,  .851,  .851}2.94 & \cellcolor[rgb]{ .851,  .851,  .851}6.47 & \cellcolor[rgb]{ .851,  .851,  .851}11.31 & \cellcolor[rgb]{ .851,  .851,  .851}9.28 & \cellcolor[rgb]{ .851,  .851,  .851}24.90 & \cellcolor[rgb]{ .851,  .851,  .851}7.84 & \cellcolor[rgb]{ .851,  .851,  .851}\cellcolor{LightBlue}9.66 & 6.00  & 0.83  & 1.92  & 6.92  & 6.42  & 20.58 & 15.42 & \cellcolor{LightBlue}8.30 \\
    SOvR \cite{hayes2022online} & 46.72 & 45.96 & 43.76 & 9.33  & 10.31 & 29.21 & 17.24 & \cellcolor{LightBlue}28.93 & \cellcolor[rgb]{ .851,  .851,  .851}80.39 & \cellcolor[rgb]{ .851,  .851,  .851}63.01 & \cellcolor[rgb]{ .851,  .851,  .851}64.97 & \cellcolor[rgb]{ .851,  .851,  .851}18.69 & \cellcolor[rgb]{ .851,  .851,  .851}8.17 & \cellcolor[rgb]{ .851,  .851,  .851}54.58 & \cellcolor[rgb]{ .851,  .851,  .851}14.51 & \cellcolor[rgb]{ .851,  .851,  .851}\cellcolor{LightBlue}43.47 & 76.33 & 63.67 & 67.50 & 17.75 & 7.00  & 50.08 & 10.00 & \cellcolor{LightBlue}41.76 \\
    SQDA \cite{anagnostopoulos2012online} & 37.65 & 47.08 & 46.10 & 37.34 & 36.20 & 41.41 & 36.41 & \cellcolor{LightBlue}40.31 & \cellcolor[rgb]{ .851,  .851,  .851}93.73 & \cellcolor[rgb]{ .851,  .851,  .851}94.18 & \cellcolor[rgb]{ .851,  .851,  .851}91.70 & \cellcolor[rgb]{ .851,  .851,  .851}96.27 & \cellcolor[rgb]{ .851,  .851,  .851}94.44 & \cellcolor[rgb]{ .851,  .851,  .851}95.03 & \cellcolor[rgb]{ .851,  .851,  .851}93.40 & \cellcolor[rgb]{ .851,  .851,  .851}\cellcolor{LightBlue}94.11 & 94.67 & 96.00 & 95.50 & 96.00 & 96.92 & 96.42 & 96.42 & \cellcolor{LightBlue}95.99 \\
    \hdashline
    iCaRL \cite{rebuffi2017icarl} & 51.29 & 51.04 & 50.33 & 33.60 & 33.57 & 42.76 & 32.58 & \cellcolor{LightBlue}42.17 & \cellcolor[rgb]{ .851,  .851,  .851}58.95 & \cellcolor[rgb]{ .851,  .851,  .851}66.21 & \cellcolor[rgb]{ .851,  .851,  .851}63.53 & \cellcolor[rgb]{ .851,  .851,  .851}34.90 & \cellcolor[rgb]{ .851,  .851,  .851}40.39 & \cellcolor[rgb]{ .851,  .851,  .851}38.50 & \cellcolor[rgb]{ .851,  .851,  .851}32.16 & \cellcolor[rgb]{ .851,  .851,  .851}\cellcolor{LightBlue}47.81 & 68.83 & 77.75 & 76.50 & 47.08 & 48.75 & 51.42 & 44.08 & \cellcolor{LightBlue}59.20 \\
    iCaRL (2pc) \cite{rebuffi2017icarl} & 49.08 & 47.28 & 47.93 & 31.83 & 28.92 & 40.89 & 31.73 & \cellcolor{LightBlue}39.67 & \cellcolor[rgb]{ .851,  .851,  .851}58.04 & \cellcolor[rgb]{ .851,  .851,  .851}67.12 & \cellcolor[rgb]{ .851,  .851,  .851}64.12 & \cellcolor[rgb]{ .851,  .851,  .851}37.25 & \cellcolor[rgb]{ .851,  .851,  .851}37.65 & \cellcolor[rgb]{ .851,  .851,  .851}38.69 & \cellcolor[rgb]{ .851,  .851,  .851}32.88 & \cellcolor[rgb]{ .851,  .851,  .851}\cellcolor{LightBlue}47.96 & 69.33 & 78.33 & 76.33 & 46.17 & 49.83 & 50.50 & 45.33 & \cellcolor{LightBlue}59.40 \\
    \midrule
    NCM/CBCL \cite{mensink2013distance,ayub2020cognitively} & 50.59 & 48.31 & 47.68 & 34.58 & 32.20 & 40.57 & 34.28 & \cellcolor{LightBlue}41.17 & \cellcolor[rgb]{ .851,  .851,  .851}94.90 & \cellcolor[rgb]{ .851,  .851,  .851}93.53 & \cellcolor[rgb]{ .851,  .851,  .851}93.66 & \cellcolor[rgb]{ .851,  .851,  .851}91.96 & \cellcolor[rgb]{ .851,  .851,  .851}92.22 & \cellcolor[rgb]{ .851,  .851,  .851}94.84 & \cellcolor[rgb]{ .851,  .851,  .851}91.70 & \cellcolor[rgb]{ .851,  .851,  .851}\cellcolor{LightBlue}93.26 & 93.83 & 94.42 & 94.33 & 93.58 & 96.17 & 96.67 & 94.17 & \cellcolor{LightBlue}94.74 \\
    RobOCLe$_{\mathrm{NCM}}$ (ours) & \textbf{55.67} & \textbf{54.84} & \textbf{54.89} & \textbf{36.39} & \textbf{32.74} & \textbf{41.23} & \textbf{35.12} & \cellcolor{LightBlue}\textbf{44.41} & \cellcolor[rgb]{ .851,  .851,  .851}\textbf{95.90} & \cellcolor[rgb]{ .851,  .851,  .851}\textbf{95.37} & \cellcolor[rgb]{ .851,  .851,  .851}\textbf{94.85} & \cellcolor[rgb]{ .851,  .851,  .851}\textbf{94.33} & \cellcolor[rgb]{ .851,  .851,  .851}\textbf{93.91} & \cellcolor[rgb]{ .851,  .851,  .851}\textbf{95.70} & \cellcolor[rgb]{ .851,  .851,  .851}\textbf{92.99} & \cellcolor[rgb]{ .851,  .851,  .851}\cellcolor{LightBlue}\textbf{94.72} & \textbf{95.86} & \textbf{96.81} & \textbf{96.51} & \textbf{95.58} & \textbf{97.00} & \textbf{97.69} & \textbf{95.71} & \cellcolor{LightBlue}\textbf{96.45} \\
    
    ($\Delta_R$ [\%]) & \scriptsize\color{OliveGreen}(+10.3) & \scriptsize\color{OliveGreen}(+12.6) & \scriptsize\color{OliveGreen}(+13.8) & \scriptsize\color{OliveGreen}(+2.8)  & \scriptsize\color{OliveGreen}(+0.8)  & \scriptsize\color{OliveGreen}(+1.1)  & \scriptsize\color{OliveGreen}(+1.3)  & \cellcolor{LightBlue}\scriptsize\color{OliveGreen}(+5.5)  & \cellcolor[rgb]{ .851,  .851,  .851}\scriptsize\color{OliveGreen}(+19.7) & \cellcolor[rgb]{ .851,  .851,  .851}\scriptsize\color{OliveGreen}(+28.5) & \cellcolor[rgb]{ .851,  .851,  .851}\scriptsize\color{OliveGreen}(+18.8) & \cellcolor[rgb]{ .851,  .851,  .851}\scriptsize\color{OliveGreen}(+29.5) & \cellcolor[rgb]{ .851,  .851,  .851}\scriptsize\color{OliveGreen}(+21.6) & \cellcolor[rgb]{ .851,  .851,  .851}\scriptsize\color{OliveGreen}(+16.) & \cellcolor[rgb]{ .851,  .851,  .851}\scriptsize\color{OliveGreen}(+15.6) & \cellcolor[rgb]{ .851,  .851,  .851}\scriptsize\color{OliveGreen}\cellcolor{LightBlue}(+21.7) & \scriptsize\color{OliveGreen}(+32.9) & \scriptsize\color{OliveGreen}(+42.8) & \scriptsize\color{OliveGreen}(+38.5) & \scriptsize\color{OliveGreen}(+31.2) & \scriptsize\color{OliveGreen}(+21.7) & \scriptsize\color{OliveGreen}(+30.6) & \scriptsize\color{OliveGreen}(+26.4) & \cellcolor{LightBlue}\scriptsize\color{OliveGreen}(+32.6) \\
    
    \midrule
    SLDA \cite{hayes2020lifelong} & 50.26 & 49.91 & 50.41 & 43.96 & 41.50 & \textbf{45.51} & 42.93 & \cellcolor{LightBlue}46.35 & \cellcolor[rgb]{ .851,  .851,  .851}94.38 & \cellcolor[rgb]{ .851,  .851,  .851}94.77 & \cellcolor[rgb]{ .851,  .851,  .851}93.99 & \cellcolor[rgb]{ .851,  .851,  .851}96.34 & \cellcolor[rgb]{ .851,  .851,  .851}95.62 & \cellcolor[rgb]{ .851,  .851,  .851}94.64 & \cellcolor[rgb]{ .851,  .851,  .851}\textbf{95.23} & \cellcolor[rgb]{ .851,  .851,  .851}\cellcolor{LightBlue}95.00 & 97.00 & 97.92 & 96.25 & 99.00 & 98.17 & 98.33 & 98.67 & \cellcolor{LightBlue}97.90 \\
    RobOCLe$_{\mathrm{SLDA}}$ (ours) & \textbf{51.33} & \textbf{51.44} & \textbf{52.42} & \textbf{44.73} & \textbf{42.86} & 45.22 & \textbf{43.07} & \cellcolor{LightBlue}\textbf{47.29} & \cellcolor[rgb]{ .851,  .851,  .851}\textbf{96.12} & \cellcolor[rgb]{ .851,  .851,  .851}\textbf{96.98} & \cellcolor[rgb]{ .851,  .851,  .851}\textbf{95.84} & \cellcolor[rgb]{ .851,  .851,  .851}\textbf{96.96} & \cellcolor[rgb]{ .851,  .851,  .851}\textbf{95.80} & \cellcolor[rgb]{ .851,  .851,  .851}\textbf{95.38} & \cellcolor[rgb]{ .851,  .851,  .851}94.73 & \cellcolor[rgb]{ .851,  .851,  .851}\cellcolor{LightBlue}\textbf{95.97} & \textbf{97.42} & \textbf{98.33} & \textbf{97.47} & \textbf{99.18} & \textbf{98.31} & \textbf{98.40} & \textbf{98.78} & \cellcolor{LightBlue}\textbf{98.27} \\
    ($\Delta_R$ [\%]) & \scriptsize\color{OliveGreen}(+2.1)  & \scriptsize\color{OliveGreen}(+3.1)  & \scriptsize\color{OliveGreen}(+4.1)  & \scriptsize\color{OliveGreen}(+1.4)  & \scriptsize\color{OliveGreen}(+2.3)  & \scriptsize\color{BrickRed}(-0.5) & \scriptsize\color{OliveGreen}(+0.2)  & \cellcolor{LightBlue}\scriptsize\color{OliveGreen}(+1.8)  & \cellcolor[rgb]{ .851,  .851,  .851}\scriptsize\color{OliveGreen}(+31.0) & \cellcolor[rgb]{ .851,  .851,  .851}\scriptsize\color{OliveGreen}(+42.2) & \cellcolor[rgb]{ .851,  .851,  .851}\scriptsize\color{OliveGreen}(+30.8) & \cellcolor[rgb]{ .851,  .851,  .851}\scriptsize\color{OliveGreen}(+17.0) & \cellcolor[rgb]{ .851,  .851,  .851}\scriptsize\color{OliveGreen}(+4.0) & \cellcolor[rgb]{ .851,  .851,  .851}\scriptsize\color{OliveGreen}(+13.7) & \cellcolor[rgb]{ .851,  .851,  .851}\scriptsize\color{BrickRed}(-10.5) & \cellcolor[rgb]{ .851,  .851,  .851}\cellcolor{LightBlue}\scriptsize\color{OliveGreen}(+19.5) & \scriptsize\color{OliveGreen}(+13.9) & \scriptsize\color{OliveGreen}(+20.0) & \scriptsize\color{OliveGreen}(+32.6) & \scriptsize\color{OliveGreen}(+17.9) & \scriptsize\color{OliveGreen}(+8.0) & \scriptsize\color{OliveGreen}(+4.3)  & \scriptsize\color{OliveGreen}(+8.4)  & \cellcolor{LightBlue}\scriptsize\color{OliveGreen}(+17.5) \\
    \bottomrule
    \end{tabular}%
    }
  \label{tab:offdomain}%
   \vspace{-5pt}
\end{table*}%

We evaluate our RobOCLe in multiple scenarios.

\noindent\textbf{Same-domain data (OpenLORIS).}
In the first analyses, we consider $16$ backbones, including both CNN-based (2x MobileNets-V3 \cite{howard2019searching}, 2x EfficientNets \cite{tan2019efficientnet}, and 5x ResNets \cite{he2016deep}) and transformer-based (3x Swins \cite{liu2021swin} and 4x ViTs \cite{dosovitskiyimage}) of variable sizes. These are commonly used baselines and allow us to examine variability of results depending on model size and learned visual features.
All architectures have been pre-trained
on the ImageNet \cite{krizhevsky2017imagenet} dataset. 

Accuracy is reported in Tab.~\ref{tab:indomain}.
Most na\"ive OCL methods (\ie, PRCPT, SNB, SOvR) are unable to outperform FT, since, in this case, data is sufficient and correlated, with no severe change of domains/objects at test time.
Replay data (\ie, iCaRL) shows benefits, however, contradicts the important OCL assumption on data availability: robots should not store data due to privacy and storage constraints.

SLDA outperforms all other competitors thanks to the Gaussian modeling of features, while its extension (SQDA) can achieve competitive performance only on certain backbones: we argue that this is because SQDA cannot create reliable class-wise correlation matrices, especially for larger networks. 
RobOCLe brings significant improvements with respect to the best approach (SLDA), reaching an outstanding average accuracy of $99.21\%$ with a room-aware relative gain ($\Delta_R$) of $65.8\%$ \rev{over it}.
This first analysis justifies the robustness of RobOCLe to multiple backbones on same-domain test data when sufficient training data is available. This experimental case is the main benchmark employed in the existing literature \cite{mai2022}; however, we argue that it may often not be the case in practice, as we discuss next.

\begin{figure}[tb]
    \centering
    \includegraphics[trim=0.2cm 0.25cm 0.2cm 0.2cm, clip, width=0.95\linewidth]{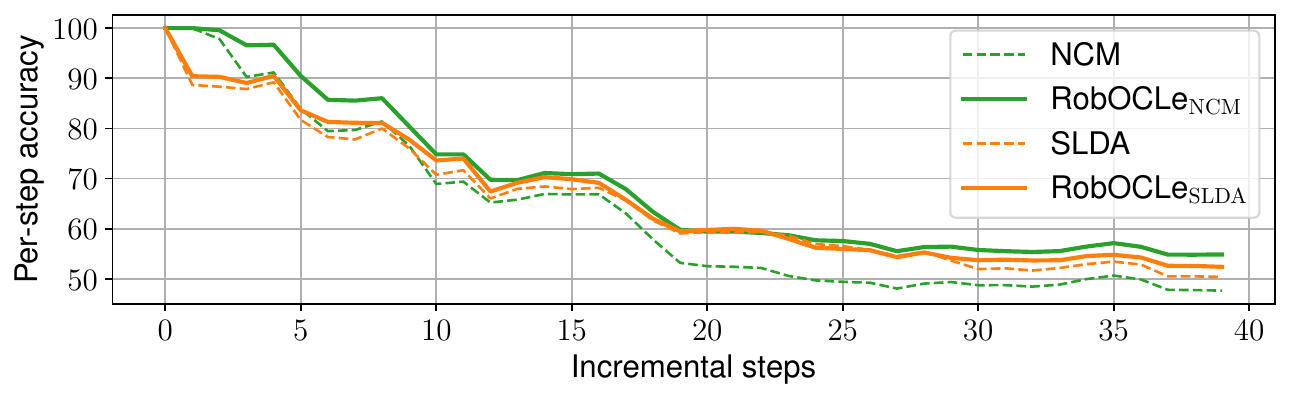}
    \vspace{-7pt}
    \caption{Per-step Acc of RN152 on OpenLORIS-small. Average gains of RobOCLe over NCM and SLDA are $5.81\%$ and $1.20\%$, respectively.}
    \vspace{-18pt}
\label{fig:perstep}
\end{figure}

\begin{table*}[t]
  \setlength{\tabcolsep}{1.5pt}
  \centering
  \caption{Accuracy on other-domain data generated via controlled augmentations on the OpenLORIS with ResNet152 and ViT-L16. \\\hl{Results highlighted in yellow}: same augmentations between train and test. tr: train, te:test.}
   \vspace{-8pt}
    \resizebox{\linewidth}{!}{%
    \begin{tabular}{c|l|cccccc|ccccccc||c|l|cccccc|ccccccc}
    \toprule
          &       & \multicolumn{6}{c|}{NCM}              & \multicolumn{7}{c||}{RobOCLe$_{\mathrm{NCM}}$ (ours)} &       &       & \multicolumn{6}{c|}{NCM}              & \multicolumn{7}{c}{RobOCLe$_{\mathrm{NCM}}$ (ours)} \\
          \cmidrule(lr){3-8}\cmidrule(lr){9-15}\cmidrule(lr){18-23}\cmidrule(lr){24-30}
    \multirow{12}[0]{*}{\begin{sideways}ResNet152\end{sideways}} & \multicolumn{1}{c|}{\textbf{tr$\downarrow$ \ te$\rightarrow$}} & \textbf{clean} & \textbf{illum} & \textbf{geom} & \textbf{noise} & \textbf{all} & \cellcolor[rgb]{ .867,  .922,  .969}\textbf{Avg OD} & \textbf{clean} & \textbf{illum} & \textbf{geom} & \textbf{noise} & \textbf{all} & \cellcolor[rgb]{ .867,  .922,  .969}\textbf{Avg OD} & \cellcolor[rgb]{ .867,  .922,  .969}$\Delta_R^{\mathrm{OD}}$ & \multirow{12}[0]{*}{\begin{sideways}ViT-L16\end{sideways}} & \multicolumn{1}{c|}{\textbf{tr$\downarrow$ \ te$\rightarrow$}} & \textbf{clean} & \textbf{illum} & \textbf{geom} & \textbf{noise} & \textbf{all} & \cellcolor[rgb]{ .867,  .922,  .969}\textbf{Avg OD} & \textbf{clean} & \textbf{illum} & \textbf{geom} & \textbf{noise} & \textbf{all} & \cellcolor[rgb]{ .867,  .922,  .969}\textbf{Avg OD} & \cellcolor[rgb]{ .867,  .922,  .969}$\Delta_R^{\mathrm{OD}}$ \\
    \midrule
          & \textbf{clean} & \cellcolor[rgb]{ 1,  .949,  .8}84.12 & 34.75 & 77.37 & 60.82 & 13.40 & \cellcolor[rgb]{ .867,  .922,  .969}46.59 & \cellcolor[rgb]{ 1,  .949,  .8}\textbf{88.15} & 41.26 & 82.10 & 66.16 & 18.16 & \cellcolor[rgb]{ .867,  .922,  .969}\textbf{51.92} & \cellcolor[rgb]{ .867,  .922,  .969}\color{OliveGreen}(+10.0)  &       & \textbf{clean} & \cellcolor[rgb]{ 1,  .949,  .8}79.47 & 35.11 & 73.43 & 66.57 & 19.49 & \cellcolor[rgb]{ .867,  .922,  .969}48.65 & \cellcolor[rgb]{ 1,  .949,  .8}\textbf{80.31} & 33.13 & 73.69 & 68.38 & 19.74 & \cellcolor[rgb]{ .867,  .922,  .969}\textbf{48.73} & \cellcolor[rgb]{ .867,  .922,  .969}\color{OliveGreen}(+0.2) \\
          & \textbf{illum} & 56.91 & \cellcolor[rgb]{ 1,  .949,  .8}53.91 & 54.77 & 45.75 & 30.29 & \cellcolor[rgb]{ .867,  .922,  .969}46.93 & 62.65 & \cellcolor[rgb]{ 1,  .949,  .8}\textbf{56.83} & 59.92 & 51.26 & 31.91 & \cellcolor[rgb]{ .867,  .922,  .969}\textbf{51.44} & \cellcolor[rgb]{ .867,  .922,  .969}\color{OliveGreen}(+8.5)   &       & \textbf{illum} & 30.78 & \cellcolor[rgb]{ 1,  .949,  .8}40.58 & 33.00 & 29.12 & 30.00 & \cellcolor[rgb]{ .867,  .922,  .969}30.73 & 30.82 & \cellcolor[rgb]{ 1,  .949,  .8}\textbf{35.00} & 35.17 & 30.65 & 31.22 & \cellcolor[rgb]{ .867,  .922,  .969}\textbf{31.96} & \cellcolor[rgb]{ .867,  .922,  .969}\color{OliveGreen}(+1.8) \\
          & \textbf{geom} & 82.01 & 40.49 & \cellcolor[rgb]{ 1,  .949,  .8}81.07 & 64.05 & 19.67 & \cellcolor[rgb]{ .867,  .922,  .969}51.55 & 86.21 & 45.11 & \cellcolor[rgb]{ 1,  .949,  .8}\textbf{84.39} & 67.92 & 23.90 & \cellcolor[rgb]{ .867,  .922,  .969}\textbf{55.79} & \cellcolor[rgb]{ .867,  .922,  .969}\color{OliveGreen}(+8.7)   &       & \textbf{geom} & 78.01 & 36.62 & \cellcolor[rgb]{ 1,  .949,  .8}79.19 & 68.73 & 23.31 & \cellcolor[rgb]{ .867,  .922,  .969}51.67 & 78.70 & 37.05 & \cellcolor[rgb]{ 1,  .949,  .8}\textbf{78.64} & 68.87 & 24.32 & \cellcolor[rgb]{ .867,  .922,  .969}\textbf{52.24} & \cellcolor[rgb]{ .867,  .922,  .969}\color{OliveGreen}(+1.2) \\
          & \textbf{noise} & 77.27 & 35.60 & 71.99 & \cellcolor[rgb]{ 1,  .949,  .8}74.31 & 20.08 & \cellcolor[rgb]{ .867,  .922,  .969}51.23 & 82.78 & 43.99 & 78.67 & \cellcolor[rgb]{ 1,  .949,  .8}\textbf{78.82} & 24.84 & \cellcolor[rgb]{ .867,  .922,  .969}\textbf{57.57} & \cellcolor[rgb]{ .867,  .922,  .969}\color{OliveGreen}(+13.0)  &       & \textbf{noise} & 73.67 & 36.37 & 66.27 & \cellcolor[rgb]{ 1,  .949,  .8}75.64 & 25.41 & \cellcolor[rgb]{ .867,  .922,  .969}50.43 & 73.84 & 36.58 & 68.55 & \cellcolor[rgb]{ 1,  .949,  .8}\textbf{76.51} & 25.56 & \cellcolor[rgb]{ .867,  .922,  .969}\textbf{51.13} & \cellcolor[rgb]{ .867,  .922,  .969}\color{OliveGreen}(+1.4) \\
          & \textbf{all} & 40.45 & 43.49 & 38.22 & 32.35 & \cellcolor[rgb]{ 1,  .949,  .8}35.02 & \cellcolor[rgb]{ .867,  .922,  .969}38.63 & 43.07 & 44.33 & 41.64 & 35.95 & \cellcolor[rgb]{ 1,  .949,  .8}\textbf{37.33} & \cellcolor[rgb]{ .867,  .922,  .969}\textbf{41.25} & \cellcolor[rgb]{ .867,  .922,  .969}\color{OliveGreen}(+4.3)   &       & \textbf{all} & 15.32 & 32.27 & 21.58 & 15.25 & \cellcolor[rgb]{ 1,  .949,  .8}31.79 & \cellcolor[rgb]{ .867,  .922,  .969}21.11 & 17.05 & 32.26 & 20.98 & 18.21 & \cellcolor[rgb]{ 1,  .949,  .8}\textbf{24.77} & \cellcolor[rgb]{ .867,  .922,  .969}\textbf{22.13} & \cellcolor[rgb]{ .867,  .922,  .969}\color{OliveGreen}(+1.3) \\
          \cmidrule{2-15}\cmidrule{17-30}
          &       & \multicolumn{6}{c|}{SLDA}             & \multicolumn{6}{c}{RobOCLe$_{\mathrm{SLDA}}$ (ours)} &       &       &       & \multicolumn{6}{c|}{SLDA}             & \multicolumn{7}{c}{RobOCLe$_{\mathrm{SLDA}}$ (ours)} \\
          \cmidrule{2-15}\cmidrule{17-30}
          & \textbf{clean} & \cellcolor[rgb]{ 1,  .949,  .8}99.13 & 47.19 & 92.55 & 77.09 & 19.74 & \cellcolor[rgb]{ .867,  .922,  .969}59.14 & \cellcolor[rgb]{ 1,  .949,  .8}\textbf{99.78} & 48.69 & 93.31 & 80.17 & 20.32 & \cellcolor[rgb]{ .867,  .922,  .969}\textbf{60.63} & \cellcolor[rgb]{ .867,  .922,  .969}\color{OliveGreen}(+3.6)   &       & \textbf{clean} & \cellcolor[rgb]{ 1,  .949,  .8}98.69 & 38.30 & 92.94 & 90.09 & 21.49 & \cellcolor[rgb]{ .867,  .922,  .969}60.71 & \cellcolor[rgb]{ 1,  .949,  .8}\textbf{99.73} & 40.69 & 95.04 & 93.03 & 22.53 & \cellcolor[rgb]{ .867,  .922,  .969}\textbf{62.82} & \cellcolor[rgb]{ .867,  .922,  .969}\color{OliveGreen}(+5.4) \\
          & \textbf{illum} & 86.68 & \cellcolor[rgb]{ 1,  .949,  .8}79.65 & 77.76 & 61.88 & 39.41 & \cellcolor[rgb]{ .867,  .922,  .969}66.43 & 89.02 & \cellcolor[rgb]{ 1,  .949,  .8}\textbf{82.21} & 79.29 & 64.58 & 40.89 & \cellcolor[rgb]{ .867,  .922,  .969}\textbf{68.45} & \cellcolor[rgb]{ .867,  .922,  .969}\color{OliveGreen}(+6.0)   &       & \textbf{illum} & 83.68 & \cellcolor[rgb]{ 1,  .949,  .8}77.13 & 72.86 & 72.39 & 50.58 & \cellcolor[rgb]{ .867,  .922,  .969}69.88 & 86.13 & \cellcolor[rgb]{ 1,  .949,  .8}\textbf{80.65} & 77.00 & 76.19 & 52.56 & \cellcolor[rgb]{ .867,  .922,  .969}\textbf{72.97} & \cellcolor[rgb]{ .867,  .922,  .969}\color{OliveGreen}(+10.3) \\
          & \textbf{geom} & 98.05 & 49.50 & \cellcolor[rgb]{ 1,  .949,  .8}97.53 & 78.08 & 22.59 & \cellcolor[rgb]{ .867,  .922,  .969}62.06 & 98.81 & 51.32 & \cellcolor[rgb]{ 1,  .949,  .8}\textbf{98.40} & 78.05 & 23.12 & \cellcolor[rgb]{ .867,  .922,  .969}\textbf{62.83} & \cellcolor[rgb]{ .867,  .922,  .969}\color{OliveGreen}(+2.0)   &       & \textbf{geom} & 97.72 & 46.72 & \cellcolor[rgb]{ 1,  .949,  .8}96.86 & 89.44 & 27.98 & \cellcolor[rgb]{ .867,  .922,  .969}65.46 & 98.74 & 45.87 & \cellcolor[rgb]{ 1,  .949,  .8}\textbf{98.10} & 92.25 & 28.08 & \cellcolor[rgb]{ .867,  .922,  .969}\textbf{66.23} & \cellcolor[rgb]{ .867,  .922,  .969}\color{OliveGreen}(+2.2) \\
          & \textbf{noise} & 97.81 & 50.14 & 90.12 & \cellcolor[rgb]{ 1,  .949,  .8}96.83 & 28.12 & \cellcolor[rgb]{ .867,  .922,  .969}66.55 & 98.72 & 51.82 & 92.85 & \cellcolor[rgb]{ 1,  .949,  .8}\textbf{98.12} & 28.57 & \cellcolor[rgb]{ .867,  .922,  .969}\textbf{67.99} & \cellcolor[rgb]{ .867,  .922,  .969}\color{OliveGreen}(+4.3)   &       & \textbf{noise} & 97.42 & 40.08 & 89.87 & \cellcolor[rgb]{ 1,  .949,  .8}97.19 & 26.21 & \cellcolor[rgb]{ .867,  .922,  .969}63.40 & 98.82 & 41.90 & 93.41 & \cellcolor[rgb]{ 1,  .949,  .8}\textbf{98.68} & 26.69 & \cellcolor[rgb]{ .867,  .922,  .969}\textbf{65.21} & \cellcolor[rgb]{ .867,  .922,  .969}\color{OliveGreen}(+4.9) \\
          & \textbf{all} & 70.16 & 68.79 & 70.85 & 66.35 & \cellcolor[rgb]{ 1,  .949,  .8}62.60 & \cellcolor[rgb]{ .867,  .922,  .969}69.04 & 74.78 & 71.70 & 74.88 & 68.73 & \cellcolor[rgb]{ 1,  .949,  .8}\textbf{65.31} & \cellcolor[rgb]{ .867,  .922,  .969}\textbf{72.52} & \cellcolor[rgb]{ .867,  .922,  .969}\color{OliveGreen}(+11.3)  &       & \textbf{all} & 71.27 & 68.15 & 66.03 & 68.67 & \cellcolor[rgb]{ 1,  .949,  .8}65.09 & \cellcolor[rgb]{ .867,  .922,  .969}68.53 & 72.81 & 71.36 & 71.58 & 71.59 & \cellcolor[rgb]{ 1,  .949,  .8}\textbf{68.43} & \cellcolor[rgb]{ .867,  .922,  .969}\textbf{71.84} & \cellcolor[rgb]{ .867,  .922,  .969}\color{OliveGreen}(+10.5) \\
          \bottomrule
    \end{tabular}%
    }
    \vspace{-11pt}
  \label{tab:offdomain_controlled_openloris}%
\end{table*}%

\begin{table}[t]
\setlength{\tabcolsep}{1.5pt}
  \centering
  \caption{\rev{Accuracy on other-domain data with controlled augmentations on RN152 on the few-shot benchmarks.}}
   \vspace{-8pt}
    \resizebox{\linewidth}{!}{%
    \begin{tabular}{c|l|cccccc|ccccccc}
          \toprule
          &       & \multicolumn{6}{c|}{SLDA}             & \multicolumn{7}{c}{RobOCLe$_{\mathrm{SLDA}}$ (ours)} \\
     & \multicolumn{1}{c|}{\textbf{tr$\downarrow$ \ te$\rightarrow$}} & \textbf{clean} & \textbf{illum} & \textbf{geom} & \textbf{noise} & \textbf{all} & \cellcolor[rgb]{ .867,  .922,  .969}\textbf{Avg$_\mathrm{OD}$} & \textbf{clean} & \textbf{illum} & \textbf{geom} & \textbf{noise} & \textbf{all} & \cellcolor[rgb]{ .867,  .922,  .969}\textbf{Avg$_\mathrm{OD}$} & \multicolumn{1}{c}{\cellcolor[rgb]{ .867,  .922,  .969}$\Delta_{R}^{\mathrm{OD}}$} \\
    \midrule
          \multirow{5}[0]{*}{\begin{sideways}\scriptsize  OLORIS-small\end{sideways}} & \textbf{clean} & \cellcolor[rgb]{ 1,  .949,  .8}50.41 & 20.06 & 44.77 & 32.80 & 10.42 & \cellcolor[rgb]{ .867,  .922,  .969}27.01 & \cellcolor[rgb]{ 1,  .949,  .8}\textbf{52.42} & 20.23 & 44.27 & 33.56 & 10.29 & \cellcolor[rgb]{ .867,  .922,  .969}\textbf{27.09} & \cellcolor[rgb]{ .867,  .922,  .969}\color{OliveGreen}(+0.1) \\
          & \textbf{illum} & 30.24 & \cellcolor[rgb]{ 1,  .949,  .8}27.71 & 20.88 & 19.28 & 12.99 & \cellcolor[rgb]{ .867,  .922,  .969}20.85 & 32.36 & \cellcolor[rgb]{ 1,  .949,  .8}\textbf{28.75} & 24.58 & 17.26 & 12.56 & \cellcolor[rgb]{ .867,  .922,  .969}\textbf{21.69} & \cellcolor[rgb]{ .867,  .922,  .969}\color{OliveGreen}(+1.1) \\
          & \textbf{geom} & 51.01 & 20.14 & \cellcolor[rgb]{ 1,  .949,  .8}49.61 & 33.76 & 10.11 & \cellcolor[rgb]{ .867,  .922,  .969}28.76 & 52.51 & 20.71 & \cellcolor[rgb]{ 1,  .949,  .8}\textbf{49.75} & 35.82 & 11.13 & \cellcolor[rgb]{ .867,  .922,  .969}\textbf{30.04} & \cellcolor[rgb]{ .867,  .922,  .969}\color{OliveGreen}(+1.8) \\
          & \textbf{noise} & 46.29 & 18.83 & 39.03 & \cellcolor[rgb]{ 1,  .949,  .8}42.31 & 11.35 & \cellcolor[rgb]{ .867,  .922,  .969}28.87 & 46.69 & 19.63 & 40.16 & \cellcolor[rgb]{ 1,  .949,  .8}\textbf{43.31} & 11.04 & \cellcolor[rgb]{ .867,  .922,  .969}\textbf{29.38} & \cellcolor[rgb]{ .867,  .922,  .969}\color{OliveGreen}(+0.7) \\
          & \textbf{all} & 23.63 & 26.08 & 22.27 & 21.40 & \cellcolor[rgb]{ 1,  .949,  .8}24.49 & \cellcolor[rgb]{ .867,  .922,  .969}23.34 & 23.85 & 28.75 & 22.88 & 23.73 & \cellcolor[rgb]{ 1,  .949,  .8}\textbf{25.80} & \cellcolor[rgb]{ .867,  .922,  .969}\textbf{24.81} & \cellcolor[rgb]{ .867,  .922,  .969}\color{OliveGreen}(+1.9) \\
          \midrule
    \multirow{5}[1]{*}{\begin{sideways}\scriptsize FSIOL310-5s\end{sideways}} & \textbf{clean} & \cellcolor[rgb]{ 1,  .949,  .8}93.99 & 36.86 & 78.89 & 58.50 & 26.08 & \cellcolor[rgb]{ .867,  .922,  .969}50.08 & \cellcolor[rgb]{ 1,  .949,  .8}\textbf{95.82} & 42.81 & 80.20 & 62.88 & 27.97 & \cellcolor[rgb]{ .867,  .922,  .969}\textbf{53.46} & \cellcolor[rgb]{ .867,  .922,  .969}\color{OliveGreen}(+6.8) \\
          & \textbf{illum} & 41.50 & \cellcolor[rgb]{ 1,  .949,  .8}60.33 & 39.48 & 33.40 & 21.11 & \cellcolor[rgb]{ .867,  .922,  .969}33.87 & 41.70 & \cellcolor[rgb]{ 1,  .949,  .8}\textbf{62.48} & 40.82 & 32.52 & 22.88 & \cellcolor[rgb]{ .867,  .922,  .969}\textbf{34.48} & \cellcolor[rgb]{ .867,  .922,  .969}\color{OliveGreen}(+0.9) \\
          & \textbf{geom} & 79.74 & 34.71 & \cellcolor[rgb]{ 1,  .949,  .8}83.59 & 40.98 & 26.54 & \cellcolor[rgb]{ .867,  .922,  .969}45.49 & 79.93 & 43.53 & \cellcolor[rgb]{ 1,  .949,  .8}\textbf{84.38} & 43.53 & 30.07 & \cellcolor[rgb]{ .867,  .922,  .969}\textbf{49.26} & \cellcolor[rgb]{ .867,  .922,  .969}\color{OliveGreen}(+6.9) \\
          & \textbf{noise} & 72.55 & 34.44 & 52.68 & \cellcolor[rgb]{ 1,  .949,  .8}83.66 & 24.12 & \cellcolor[rgb]{ .867,  .922,  .969}45.95 & 75.69 & 38.10 & 58.56 & \cellcolor[rgb]{ 1,  .949,  .8}\textbf{85.36} & 25.69 & \cellcolor[rgb]{ .867,  .922,  .969}\textbf{49.51} & \cellcolor[rgb]{ .867,  .922,  .969}\color{OliveGreen}(+6.6) \\
          & \textbf{all} & 24.25 & 25.49 & 21.31 & 17.39 & \cellcolor[rgb]{ 1,  .949,  .8}19.61 & \cellcolor[rgb]{ .867,  .922,  .969}22.11 & 26.27 & 26.86 & 19.67 & 20.59 & \cellcolor[rgb]{ 1,  .949,  .8}\textbf{22.03} & \cellcolor[rgb]{ .867,  .922,  .969}\textbf{23.35} & \cellcolor[rgb]{ .867,  .922,  .969}\color{OliveGreen}(+1.6) \\
    \midrule
    \multirow{5}[1]{*}{\begin{sideways}\scriptsize FSIOL310-10s\end{sideways}} & \textbf{clean} & \cellcolor[rgb]{ 1,  .949,  .8}96.25 & 40.17 & 85.50 & 62.75 & 29.33 & \cellcolor[rgb]{ .867,  .922,  .969}54.44 & \cellcolor[rgb]{ 1,  .949,  .8}\textbf{97.58} & 49.08 & 87.25 & 66.75 & 32.92 & \cellcolor[rgb]{ .867,  .922,  .969}\textbf{59.00} & \cellcolor[rgb]{ .867,  .922,  .969}\color{OliveGreen}(+10.0) \\
          & \textbf{illum} & 50.33 & \cellcolor[rgb]{ 1,  .949,  .8}68.83 & 39.08 & 36.08 & 28.08 & \cellcolor[rgb]{ .867,  .922,  .969}38.40 & 53.50 & \cellcolor[rgb]{ 1,  .949,  .8}\textbf{69.42} & 39.67 & 39.17 & 29.50 & \cellcolor[rgb]{ .867,  .922,  .969}\textbf{40.46} & \cellcolor[rgb]{ .867,  .922,  .969}\color{OliveGreen}(+3.3) \\
          & \textbf{geom} & 88.33 & 43.25 & \cellcolor[rgb]{ 1,  .949,  .8}90.92 & 42.25 & 32.58 & \cellcolor[rgb]{ .867,  .922,  .969}51.60 & 91.08 & 49.92 & \cellcolor[rgb]{ 1,  .949,  .8}\textbf{92.17} & 52.17 & 33.50 & \cellcolor[rgb]{ .867,  .922,  .969}\textbf{56.67} & \cellcolor[rgb]{ .867,  .922,  .969}\color{OliveGreen}(+10.5) \\
          & \textbf{noise} & 81.92 & 36.58 & 61.00 & \cellcolor[rgb]{ 1,  .949,  .8}92.08 & 24.08 & \cellcolor[rgb]{ .867,  .922,  .969}50.90 & 86.67 & 44.17 & 65.83 & \cellcolor[rgb]{ 1,  .949,  .8}\textbf{94.92} & 27.50 & \cellcolor[rgb]{ .867,  .922,  .969}\textbf{56.04} & \cellcolor[rgb]{ .867,  .922,  .969}\color{OliveGreen}(+10.5) \\
          & \textbf{all} & 27.25 & 24.42 & 27.67 & 22.42 & \cellcolor[rgb]{ 1,  .949,  .8}43.17 & \cellcolor[rgb]{ .867,  .922,  .969}25.44 & 28.58 & 25.17 & 29.83 & 23.00 & \cellcolor[rgb]{ 1,  .949,  .8}\textbf{43.83} & \cellcolor[rgb]{ .867,  .922,  .969}\textbf{26.65} & \cellcolor[rgb]{ .867,  .922,  .969}\color{OliveGreen}(+1.6) \\
          \bottomrule
    \end{tabular}%
    }
  \vspace{-15pt}
  \label{tab:offdomain_controlled_otherdatasets_SLDA}%
\end{table}%

\noindent\textbf{Real other-domain few-shot data (OpenLORIS-small, F-SIOL-310).}
When deploying robots to users and updating their AI models to recognize new objects, we typically encounter two simultaneous problems: 1) users do not label a large amount of data as it is a tedious time-consuming operation, and 2) training and testing sets have different conditions in terms of, \eg, object pose and background environment. For example, users show some pictures of their pet lying on the living room carpet during daytime, while at a later stage, the dog's pose and the environment will change.

In Tab.~\ref{tab:offdomain}, we analyse this challenging scenario of FS-OCL with a different domain (other-domain) at test time. We consider three benchmarks based on OpenLORIS-small and F-SIOL-310 (with 5 and 10 shots), while we restrict the evaluation to ResNets and ViTs, being the most popular CNNs and Transformers at the time of writing (also, we confirmed their effectiveness in Tab.~\ref{tab:indomain}).

Also in this case, FT and na\"ive OCL methods show low accuracy. SQDA achieves competitive performance only in certain cases. iCaRL improves significantly classical FT, however, it shows limited results.
We observe that when data is scarce, NCM reduces the gap with respect to SLDA. 
In some cases (\eg, ResNets on OpenLORIS-small) NCM outperforms SLDA.
We argue that this reflects two properties:
1) SLDA optimizes a larger number of parameters (prototypes and covariance matrix) compared to NCM (prototypes matrix only); and
2) feature dimensionality of ResNets (\eg 2048 for RN152) is higher than that of ViTs (\eg 1024 for ViT-L32), thus increasing the size of the covariance matrix.
Therefore, on few-shot data, both NCM and SLDA can achieve competitive results.
Our RobOCLe improves both NCM and SLDA in almost every backbone and dataset, with just 2 exceptions out of 42 scenarios.
The average gain $\Delta_R$ ranges from 1.8\% to 32.6\%, with the highest gains shown in the F-SIOL-310 dataset using lower labelled samples. 
Remarkably, the gains on the other-domain datasets reported in this paragraph are even larger than the gains reported on same-domain data on OpenLORIS in Tab.~\ref{tab:indomain}.

Overall, these analyses certify the robustness of our method to both low-shot labelled training samples (outperforming competitors with as little as 5-shots) and datasets (improving on both the OpenLORIS-small and F-SIOL-310 with different objects and/or conditions seen at test time). 

In Fig.~\ref{fig:perstep}, we also show robustness to per-step accuracy. RobOCLe consistently outperforms the baselines at every step, making it particularly suitable for real-world applications where the number of classes is not defined \textit{a priori}.

\noindent\textbf{Controlled augmentations on other-domain data (OpenLORIS, OpenLORIS-small, F-SIOL-310).}
We anticipated in Sec.~\ref{sec:setup} and Fig.~\ref{fig:datasets} that the existing datasets are limited in having either partial (OpenLORIS) or no (F-SIOL-310) different conditions at test time. 
To cope with this limitation, we apply further augmentations at train and test time.

First, we perform experiments on OpenLORIS using the best CNN (RN152) and the best transformer (ViT-L16) on the two best OCL methods (NCM and SLDA) with variable train and test augmentations. Results are reported in Tab.~\ref{tab:offdomain_controlled_openloris}.
For each sub-table, we report train augmentations on the rows and test augmentation on the columns, and we compute the average other-domain accuracy (Avg OD). For each group of results (RobOCLe vs. baseline), we also report the RARG of RobOCLe in terms of Avg OD. 
On each scenario, RobOCLe shows superior results on severe and controlled synthetic augmentations when applied in training and/or in testing stages.
Employing either color, geometric or noise augmentation generally improves Avg OD. Their combination improves Avg OD when using SLDA only, as it has more capacity than NCM to capture and disentangle input-level domain variations.
When non-clean same augmentation is applied to both train and test sets, instead, we obtain lower accuracy compared to the original accuracy obtained using clean train/test sets. This is due to the tasks becoming much harder as, in the OCL setup, the agent experiences data only once. Therefore, convergence of models to learn robust representations is hindered.
Also in this case, we confirm how RobOCLe is more effective than the baseline counterpart, especially on other-domain setups.

\rev{Next, we select RN152, and analyse the effect of controlled augmentations on the few-shot datasets using SLDA in Tab.~\ref{tab:offdomain_controlled_otherdatasets_SLDA}.}
For each dataset and for each training augmentation, our RobOCLe robustly outperforms the baseline by a large margin on both same-domain and other-domain data.

\begin{table}[t]
\setlength{\tabcolsep}{1pt}
  \centering
  \caption{Multiple pooling mechanisms on OpenLORIS-small and SLDA. $\Delta_{f}$: the average feature size multiplier compared to AVG.}
  \vspace{-8pt}
  \resizebox{\linewidth}{!}{%
    \begin{tabular}{lccccccccc}
    \toprule
          & \multirow{2}[2]{*}{$\Delta_{f}$} & \multicolumn{8}{c}{\textbf{Accuracy}} \\ \cmidrule(lr){3-10}
          &       & \textbf{RN50} & \textbf{RN101} & \textbf{RN152} & \textbf{ViT-B16} & \textbf{ViT-B32} & \textbf{ViT-L16} & \textbf{ViT-L32} & \cellcolor{LightBlue}\textbf{Avg} \\
    \midrule
    AVG \cite{lecun1998gradient} & 1.0   & 50.26 & 49.91 & 50.41 & 43.96 & 41.50 & 45.51 & 42.93 & \cellcolor{LightBlue}46.35 \\
    \texttt{[CLS]} \cite{kenton2019bert} & 1.0   & -     & -     & -     & 43.10 & 40.78 & 45.57 & 41.83 & \cellcolor{LightBlue}- \\
    MAX \cite{riesenhuber1999hierarchical} & 1.0   & 50.87 & 50.01 & 50.75 & 41.27 & 38.67 & 43.23 & 39.79 & \cellcolor{LightBlue}44.94 \\
    AVGMAX \cite{monteiro2020performance} & 2.0   & 50.74 & 50.16 & 50.51 & 42.21 & 39.49 & 44.42 & 40.19 & \cellcolor{LightBlue}45.39 \\
    MIX (50\%) \cite{zhou2021mixed} & 1.0   & 50.04 & 49.48 & 50.14 & 43.07 & 40.86 & 45.15 & 40.84 & \cellcolor{LightBlue}45.65 \\
    STOCHASTIC \cite{zeiler2013stochastic} & 1.0   & 42.15 & 40.37 & 38.48 & 32.78 & 29.45 & 31.97 & 30.68 & \cellcolor{LightBlue}35.13 \\
    L2  \cite{feng2011geometric} & 1.0   & 44.12 & 43.71 & 44.09 & 35.14 & 30.88 & 36.19 & 32.27 & \cellcolor{LightBlue}38.06 \\
    L3 \cite{feng2011geometric} & 1.0   & 45.12 & 43.15 & 43.80 & 36.44 & 31.11 & 37.86 & 32.57 & \cellcolor{LightBlue}38.58 \\
    RAP (1\%) \cite{bera2020effect} & 2.8   & 50.39 & 50.14 & 50.01 & 43.11 & 41.98 & 44.48 & 42.61 & \cellcolor{LightBlue}46.10 \\
    RAP (10\%) \cite{bera2020effect} & 139.0 & 39.48 & 32.14 & 31.15 & 27.48 & 24.41 & 26.10 & 24.13 & \cellcolor{LightBlue}29.27 \\
    iSQRT-COV \cite{li2018towards} & 1.0   & 50.34 & 48.50 & 48.75 & 42.98 & 41.66 & \textbf{45.75} & 42.12 & \cellcolor{LightBlue}45.73 \\
    RobOCLe$_{\mathrm{SLDA}}$ (ours) & 3.0   & \textbf{51.33} & \textbf{51.44} & \textbf{52.42} & \textbf{44.73} & \textbf{42.86} & 45.22 & \textbf{43.07} & \cellcolor{LightBlue}\textbf{47.29} \\
    \bottomrule
    \end{tabular}%
    }
 \vspace{-5pt}
  \label{tab:ablation_pooling}%
\end{table}%

\begin{table}[t]
  \centering
  \setlength{\tabcolsep}{2.2pt}
  \caption{Additional metrics on the OpenLORIS-small and ResNet152. TTime: training time [min]. FPS: frames per sec at test time.}
 \vspace{-8pt}
  \resizebox{\linewidth}{!}{%
    \begin{tabular}{lcccccc}
    \toprule
          & \textbf{Acc} $\uparrow$ & \textbf{BwT} $\uparrow$ & \textbf{ Forg} $\downarrow$ & \textbf{Pla} $\uparrow$ & \textbf{TTime} $\downarrow$ & \textbf{FPS} $\uparrow$ \\
    \midrule
    FT    & 16.05 & 23.37 & 78.77 & 92.82 & 5.57 & 165.8 \\
    \midrule
    NCM   & 47.68 & 57.34 & 46.54 & 61.10 & 3.93  & 164.8 \\
    RobOCLe$_{\mathrm{NCM}}$ (ours) & 54.89 \scriptsize\color{OliveGreen}(+13.8\%) & 63.85 & 39.95 & 69.18 & 4.24  & 164.2 \scriptsize\color{BrickRed}(-0.36\%) \\
    \midrule
    SLDA  & 50.41 & 59.14 & 45.69 & 66.19 & 4.71  & 165.0 \\
    RobOCLe$_{\mathrm{SLDA}}$ (ours) & 52.42 \scriptsize\color{OliveGreen}(+4.1\%) & 60.75 & 43.54 & 70.38 & 7.14 & 164.2 \scriptsize\color{BrickRed}(-0.45\%) \\
    \bottomrule
    \end{tabular}%
    }
    \vspace{-16pt}
  \label{tab:ablation_additional_metrics}%
\end{table}%

\noindent\textbf{Ablation studies.}
So far, we presented evaluation of results on the basis of 1) mostly replay-free OCL methods, and 2) accuracy. Here, we complement the analyses.
In Tab.~\ref{tab:ablation_pooling}, we report accuracy of other pooling methods described in Sec.~\ref{sec:method}. We observe that \texttt{[CLS]} pooling performs slightly worse than AVG, due to the limited input data. 
Concatenating the top-k\% features (RAP) increases the feature size but shows lower accuracy compared to the proposed method: using larger feature size is not enough to improve accuracy and robustness.
All other approaches do not achieve competitive results. 
Our method slightly increases the feature size, however, it brings robust recognition results.

As we introduced in Sec.~\ref{sec:setup}, multiple metrics have been defined to characterize performance of CL agents.
Tab.~\ref{tab:ablation_additional_metrics} summarizes the main results obtained on the OpenLORIS-small.
RobOCLe improves performance with respect to baseline competitors with higher Acc (overall final performance), BwT (knowledge transfer to past classes), and lower Forg (average forgetting of all classes), by finding a better trade-off between stability and Pla (which is lower than, \eg, FT).
For completeness, FwT is always 0, since incremental steps contain disjoint classes.
FT implements the classifiers as a fully-connected layer and shows longer TTime than NCM and SLDA, which only requires  computation of distance between embedded features and class prototypes.
RobOCLe adds some computational overhead which is mostly affecting TTime, while inference FPS remains practically unchanged with a small decrease of less than 0.5\% in all cases. On the other hand, the slight increase of TTime is not a relevant concern, since training is done on few-shot data only.

\section{CONCLUSION}
\label{sec:conclusion}

In this paper, we tackled the practical task of FS-OCL targeting robust test-time object recognition for low-resource robots with limited labelled data and computational/storage capability.
We introduced RobOCLe that promotes invariance to augmentation via high order statistical moments of the embedded features of input samples. 
We proved that RobOCLe achieves robust recognition in a variety of scenarios, using several backbones, low-shot setups, per-step accuracy, and controlled train/test augmentation on both same-domain and other-domain data.
Overall, FS-OCL task in low-resource devices is far from being solved.
We hope that our problem formulation, approach and extensive comparison with previous methods will encourage future works on this direction.

\bibliographystyle{IEEEtran}
\bibliography{strings_short,umbib}

\end{document}